\documentclass[fleqn,13pt]{wlscirep}
\usepackage[utf8]{inputenc}
\usepackage[T1]{fontenc}
\usepackage{bm}
\usepackage{caption}
\usepackage{subcaption}
\usepackage{ragged2e}

\title{Local learning for stable backpropagation-free neural network training towards physical learning}

\author[1]{Yaqi Guo}
\author[1,2]{Fabian Braun}
\author[2]{Bastiaan Ketelaar}
\author[3]{Stephanie Tan}
\author[2$\dagger\star$]{Richard Norte}
\author[1$\dagger\star$]{Siddhant Kumar}

\affil[1]{Department of Materials Science and Engineering, Delft University of Technology, Mekelweg 2, Delft, 2628CD, The Netherlands}
\affil[2]{Department of Precision and Microsystems Engineering, Delft University of Technology, Mekelweg 2, Delft, 2628CD, The Netherlands}
\affil[3]{Department of Intelligent Systems, Delft University of Technology, Van Mourik Broekmanweg 6, Delft 2628XE, The Netherlands}

\affil[$\dagger$]{These authors jointly supervised this work.}
\affil[$\star$]{R.A.Norte@tudelft.nl, Sid.Kumar@tudelft.nl}


\newcommand{\boldface}[1]{\boldsymbol{#1}}  % italic (slanted)

\newcommand{\bfu}{\boldface{u}}
\newcommand{\bfv}{\boldface{v}}

\newcommand{\bfx}{\boldface{x}}
\newcommand{\bfy}{\boldface{y}}
\newcommand{\bfz}{\boldface{z}}
\newcommand{\bfA}{\boldface{A}}

\newcommand{\bfI}{\boldface{I}}

\newcommand{\bfW}{\boldface{W}}

\newcommand{\bfxi}{\boldsymbol{\xi}}

\newcommand{\bfomega}{\boldsymbol{\omega}}

\newcommand{\calG}{\mathcal{G}}

\newcommand{\calL}{\mathcal{L}}

\newcommand{\calN}{\mathcal{N}}

\newcommand{\calY}{\mathcal{Y}}

\newcommand{\partderiv}[2]{\frac{\partial #1}{\partial #2}}

\newcommand{\Rset}{\mathbb{R}}

\newlength{\boxwidth}
\def\btheorem{\begin{theorem}}
\def\etheorem{\end{theorem}}
\def\blemma{\begin{lemma}}
\def\elemma{\end{lemma}}
\def\bproposition{\begin{proposition}}
\def\eproposition{\end{proposition}}
\def\bcorollary{\begin{corollary}}
\def\ecorollary{\end{corollary}}
\def\bdefinition{\begin{definition}}
\def\edefinition{\end{definition}}
\def\bexample{\begin{example}}
\def\eexample{\end{example}}
\def\bremark{\begin{remark}}
\def\eremark{\end{remark}}

\DeclareMathOperator*{\argmax}{{arg\,max}}
\DeclareMathOperator*{\argmin}{{arg\,min}}

\newcommand{\be}{\begin{equation}}
\newcommand{\ee}{\end{equation}}
\newcommand{\beq}{\begin{eqnarray}}
\newcommand{\eeq}{\end{eqnarray}}
\newcommand{\bem}{\begin{multline}}
\newcommand{\eem}{\end{multline}}
\newcommand{\ba}{\begin{align}}
\newcommand{\ea}{\end{align}}

\usepackage{graphicx}%
\usepackage{multirow}%
\usepackage{amsmath,amssymb,amsfonts}%
\usepackage{amsthm}%
\usepackage{mathrsfs}%
\usepackage[title]{appendix}%
\usepackage{xcolor}%
\usepackage{tabularray}%
\usepackage{textcomp}%
\usepackage{manyfoot}%
\usepackage{booktabs}%
\usepackage{tabularx}%
\usepackage{makecell}%
\usepackage{algorithm}%
\usepackage{algorithmicx}%
\usepackage{algpseudocode}%
\usepackage{listings}%
\usepackage{soul}
\usepackage{bbm}

\begin{abstract}
While backpropagation and automatic differentiation have driven deep learning's success, the physical limits of chip manufacturing and rising environmental costs of deep learning motivate alternative learning paradigms such as physical neural networks. However, most existing physical neural networks still rely on digital computing for training, largely because backpropagation and automatic differentiation are difficult to realize in physical systems. We introduce FFzero, a forward-only learning framework enabling stable neural network training without backpropagation or automatic differentiation. FFzero combines layer-wise local learning, prototype-based representations, and directional-derivative-based optimization through forward evaluations only. We show that local learning is effective under forward-only optimization, where backpropagation fails. FFzero generalizes to multilayer perceptron and convolutional neural networks across classification and regression. Using a simulated photonic neural network as an example, we demonstrate that FFzero provides a viable path toward backpropagation-free in-situ physical learning.
\end{abstract}

\begin{document}

\flushbottom
\maketitle

\thispagestyle{empty}

\large

\section*{Introduction}

Artificial intelligence (AI) and its rapid development carry an escalating environmental cost as training large deep learning models incurs significant energy consumption and carbon footprints. \cite{chen2025much} Training a state-of-the-art transformer model with architecture optimization has been estimated to emit roughly five times more carbon dioxide than the full lifetime emissions of an average car including fuel use. \cite{strubell2019energy, gowda2024watt} At the same time, the number of parameters in state-of-the-art AI models, and consequently the compute required to train them, is increasing at an exponential rate, while uncertainty about the continued validity of Moore’s law to support this trend from the hardware side is rising as transistors approach atomic scales. \cite{shalf2015computing}

This over-reliance on compute scaling is beginning to push against assumptions in the von Neumann architecture. These machines were designed for sequential instruction fetches and data operations, which are not suitable for the large-scale dense linear algebra that dominates modern deep learning. \cite{ganguly2019towards, heller1978survey, kimovski2023beyond} GPUs emerged as high-throughput accelerators, but their own von Neumann bottlenecks are now reaching the limit with increasing compute scaling in AI. \cite{zidan2018future} This highlights the need for a new computing paradigm for training AI models, specifically deep neural networks.

As an alternative to von Neumann or digital computing, physical or analog computing, supported by neuromorphic inspirations, is making a resurgence after decades of dormancy. \cite{ye2025overhead, seok2024beyond, mondal2024recent} A physical system can be used as a reservoir for computing or transforming signals in a fast and efficient way. For example, a harmonic pendulum can be thought of as a simple sine-calculator. In a similar vein, physical computing leverages high-dimensional, nonlinear dynamical and physical systems,\cite{tanaka2019recent} such as mechanical, \cite{nelson2024spectral} optomechanical,\cite{kiyabu2025optomechanical} photonic,\cite{van2017advances} electrical,\cite{taniguchi2021reservoir, appeltant2011information} electromechanical,\cite{guo2024mems} and iontronic\cite{Conte2025} networks to transform input signals into rich nonlinear outputs. \cite{yan2024emerging, nakajima2020physical} In the simplest case of reservoir computing, a fixed physical reservoir is appended with a trainable linear readout layer, allowing for efficient training via classical system identification techniques. Similar to stacking linear layers and activation functions in digital neural networks, these reservoirs can be daisy-chained and interleaved with linear trainable layers to create physical analogs of neural networks. \cite{nakajima2021scalable, moon2021hierarchical}

A longstanding challenge in the development of physical learning systems is the problem of training itself, and more specifically the implementation of backpropagation. \cite{momeni2025training} Digital neural networks are typically trained by solving an optimization problem that minimizes a loss function, which is evaluated on the forward pass of each dataset sample through the network. At the core of neural network training is the backpropagation algorithm, which calculates the gradient of the loss function with respect to each trainable parameter using the chain rule of differentiation. \cite{rumelhart1986learning, lillicrap2020backpropagation} Automatic differentiation enables the efficient computation of exact gradients by storing information going forward from inputs to outputs that can be reused going backwards from outputs to inputs (i.e., backpropagation). \cite{paszke2017automatic}
 This process enables iterative weight updates via, e.g., gradient descent, progressively reducing the loss and improving the network’s performance. However, unlike in software digital computing, obtaining exact gradients for a  physical system is generally not possible. \cite{ambrogio2018equivalent, laborieux2022holomorphic} This raises a question: how can physical systems be trained without backpropagation and, more broadly, automatic differentiation?

Recent works have addressed this by creating a digital twin or surrogate simulation of the physical system and computing the gradients through that model, which is coined physics-aware training. \cite{wright2022deep} However, small discrepancies between the physical components and their digital twin can produce errors in gradient calculation, which may compound across multiple layers and ultimately destabilize training, particularly in deep networks. \cite{wright2022deep, nakajima2022physical} More critically, the scalability of physical learning would be limited by the
scalability of the classical computer hardware and the digital twin. Additionally, relying on a hybrid in-situ and in-silico approach for training is not suitable for true in-situ continual learning, where external computation may not be available. 

Recently, specialized physical systems have been demonstrated\cite{pai2023experimentally,Ashtiani2026} in which analog gradients are computed by physically propagating signals forward and backward through the system. However, these approaches are restricted to highly specific platforms, particularly photonic and optical systems where signals can propagate bidirectionally with negligible loss, and therefore do not present a general-purpose solution applicable to a wide range of physical systems.

An alternative approach is to estimate gradients only from forward evaluations (also known as zeroth-order perturbative methods), thereby avoiding backpropagation. \cite{baydin2022gradients} Each weight is perturbed individually followed by a loss evaluation, and a gradient approximated via finite differences.\cite{larson2019derivative} While feasible for small networks, this direct partial derivative estimation scales poorly for deep or wide networks with large number of parameters (e.g., a million-parameter model will require a million forward passes). Directional derivative methods\cite{spall2002implementation,yin2001random}  further approximate the gradient by perturbing all the weights simultaneously in a random direction, thus requiring only two forward passes. Bandyopadhyay et al.\cite{bandyopadhyay2024single} applied  this strategy to train an on-chip photonic deep neural network. However, approximating gradients with directional derivatives can introduce significant errors that compound across layers, destabilizing training and degrading performance in large networks (as later demonstrated in our work).

Other approaches include Hebbian learning and contrastive Hebbian updates,\cite{journe2022hebbian} equilibrium-based learning,\cite{laydevant2024training} and direct feedback alignment,\cite{nokland2016direct}  which have all shown promise for physical computing. \cite{nakajima2022physical,momeni2025training} However, each approach presents its own challenges:\cite{Chen2025} Hebbian and equilibrium-based learning require physically difficult-to-implement bidirectional systems, while direct feedback alignment exhibits limited competitiveness in deep and convolutional networks.

Hinton introduced the forward-forward (FF) learning approach as an alternative to backpropagation. \cite{hinton2022forward} The FF method involves two forward passes: one with positive data (correct labels) and another with negative data (incorrect labels). Instead of computing gradients and propagating errors backward, FF updates network parameters by comparing the ``goodness'' between these passes. Goodness is defined as maximizing the difference in activations between positive and negative data at each intermediate layer, enabling ``local learning'' where each layer is trained independently, bypassing the need for backpropagation across the entire network. This makes the approach particularly relevant for analog or physical learning systems, where black-box physical nonlinearities can be interleaved with trainable linear layers without requiring gradient propagation through the physical components. However, FF's in-silico\cite{hinton2022forward,Padmani2025,Scodellaro2025} or physical implementations\cite{momeni2023backpropagation} rely on either automatic differentiation via CPU/GPU hooks or digital twins to compute layer-wise local gradients, which limits its suitability for true in-situ physical learning. 

For example, Momeini et al. \cite{momeni2023backpropagation} demonstrated application of FF to train physical learners comprising of nonlinear elements and interleaved with linear readout layers  based on acoustic, microwave, and optical systems. While FF enabled in-situ training, it still required a digital GPU to calculate local goodness and train the independent linear layers. Although FF breaks global learning into local optimizations, each local optimization remains nonlinear and requires backpropagation and gradient descent. The reliance on external CPU/GPUs
for training---even with the FF algorithm---undermines the goal of achieving scalable and true physical learning.

\begin{figure}[t]
    \centering
\includegraphics[width=\linewidth]{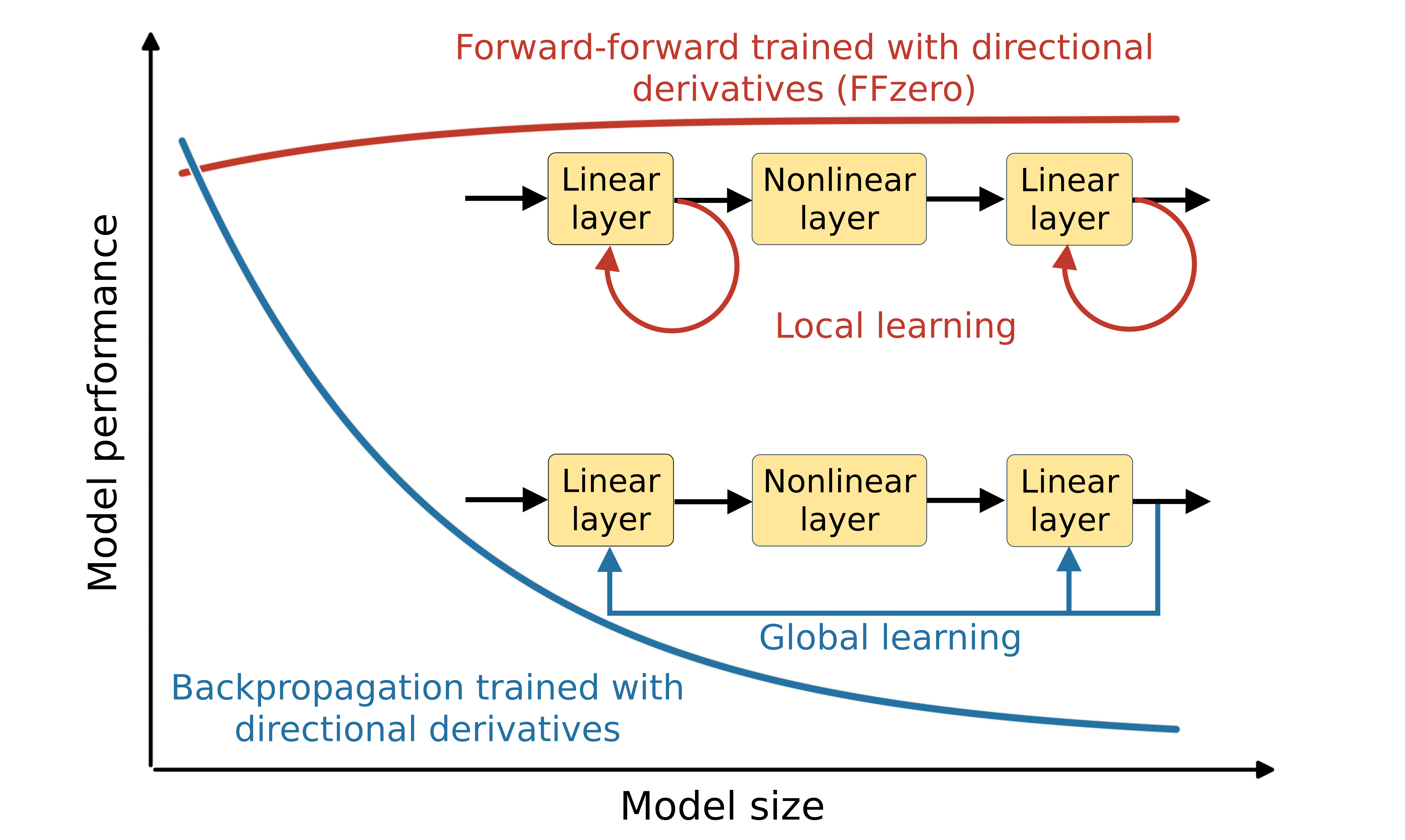}
\caption{Schematic comparison of FFzero and backpropagation performance when trained with directional derivatives. FFzero employs a local learning strategy in which each layer is updated independently, preventing gradient-estimation errors from compounding across layers. In contrast, backpropagation propagates directional-derivative estimation errors globally across layers, causing performance to degrade with increasing model size. FFzero maintains stable performance at scale while remaining compatible with in-situ physical implementation.
}
\label{fig:ff-schematic}
\end{figure}

To overcome these limitations, we introduce \textit{FFzero}, a framework that combines forward-forward local learning with zeroth-order perturbative weight updates. The central hypothesis is that perturbative methods---even with error-prone directional derivatives---can provide competitive performance in machine learning tasks in the local learning setting of forward-forward training, on the basis that gradient estimation errors do not compound across layers. We contrast FFzero with classical global learning and show that the latter is not well suited to perturbative methods and therefore is not competitive or scalable for large networks. We evaluate FFzero on feedforward and convolutional neural network architectures across diverse multiclass classification and regression benchmarks. Results show that FFzero consistently outperforms backpropagation trained with global directional derivatives, especially for deep networks, underscoring its potential for truly in-situ physical learning (see Fig.~\ref{fig:ff-schematic} for a schematic).

\section*{Results}

\subsection*{Problem setting}

To clearly illustrate the proposed FFzero algorithm, we focus on a simple two-layer multilayer perceptron (MLP) network solely for explanatory purposes, while noting that the method readily generalizes to deeper and more complex architectures, as shown later. Consider a network given by 
\begin{equation}
\underbrace{\bfz_1 = \bfW_1\bfx}_{\text{linear layer \#1}},\qquad \underbrace{\bfu=\sigma(\bfz_1)}_{\text{nonlinear activation}}, \qquad \underbrace{\bfz_2 = \bfW_2 \bfu}_{\text{linear layer \#2}},
\label{eq:simple_layers}
\end{equation}
where $\bfx$ is the input signal. $\bfW_1$ and $\bfW_2$ are weight parameter matrices for linear layers \#1 and \#2, respectively, and $\sigma(\cdot)$ represents a nonlinear activation function introduced to provide greater approximation capacity. $\bfz_1$ and $\bfz_2$ are the outputs of the linear layers, and $\bfu$ is the output of the activation function. Bias terms can be included in each linear layer but are omitted here for notational simplicity.
For examples of mechanical, optical, or photonic  realizations of such a network, we refer to refs. \cite{wright2022deep,bandyopadhyay2024single, momeni2023backpropagation,momeni2025training} In these systems, data may be encoded and decoded through the amplitude, frequency, or phase of the underlying analog signals. A loss function $\mathcal{L}$ quantifies the discrepancy between predictions $\bfz_2$ and the ground truth $\bfy_{\text{GT}}$, e.g., cross-entropy loss for classification or mean squared error loss for regression. Note that we do not assume $\dim(\bfz_2)=\dim(\bfy_\text{GT})$ to maintain generality.

In conventional backpropagation, the training objective is to minimize $\mathcal{L}$ with respect to the network weight parameters, represented as 
\begin{equation}\label{eq:bp-opt}
\bfW_1,\bfW_2 \leftarrow \argmin_{\bfW_1,\bfW_2} \calL(\bfz_2(\bfW_1,\bfW_2,\bfx),\bfy_\text{GT}).
\end{equation}
The optimization is performed in an iterative fashion wherein the weights are updated incrementally (with learning rate $\lambda>0$) as 
\be
\bfW_1 \leftarrow \bfW_1 - \lambda \frac{\partial \calL}{\partial \bfW_1} \qquad \text{and} \qquad \bfW_2 \leftarrow \bfW_2 - \lambda \frac{\partial \calL}{\partial \bfW_2}
\ee
until convergence over a sufficiently large dataset of $(\bfx, \bfy_{\text{GT}})$ pairs. Evaluating partial derivatives for training requires backpropagation through both linear and nonlinear transformations via the chain rule of differentiation. This poses significant challenges for the implementation of physical in-situ training, since the recursive nature of gradient computation is not realized and typically requires external digital computation or surrogate modeling.

\begin{figure}
    \centering
\includegraphics[width=\linewidth]{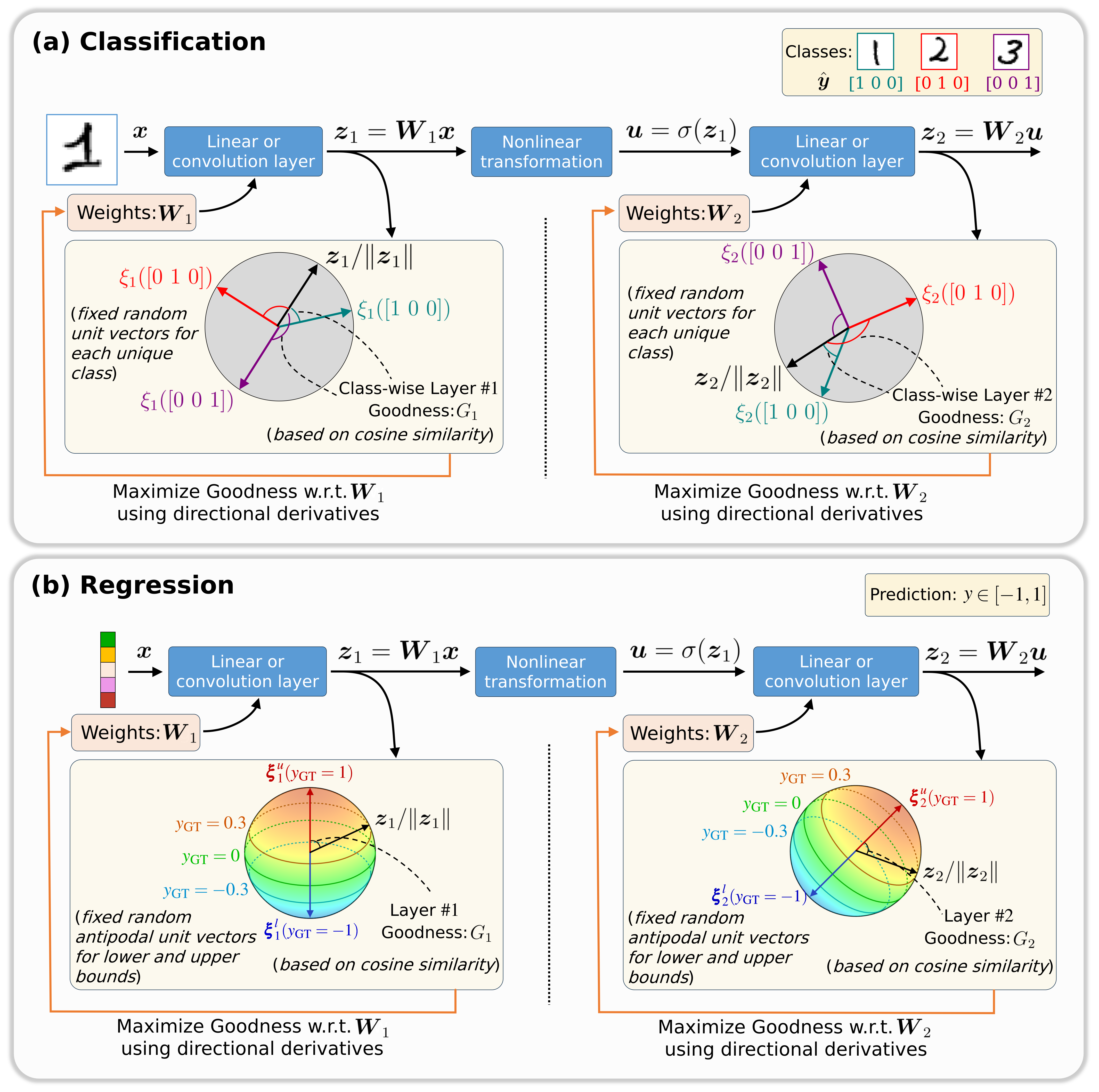}
\caption{Schematic of FFzero algorithm, illustrated for a \textbf{(a)}  3-class classification problem and \textbf{(b)} regression problem. \textbf{(a)} Each linear layer is trained locally and independently by maximizing layer-wise goodness. The cosine similarities of the layer output w.r.t. the fixed prototype vectors (denoted by $\xi[\cdot]$) are used to compute the goodness. Directional derivatives are used to perform gradient-based optimization for each layer locally. \textbf{(b)} In regression tasks, the antipodal prototypes represent the upper ($y=1$) and lower ($y=-1$) prediction bounds,
with intermediate ground-truth values encoded as interpolations between them. Each layer’s output is normalized onto a unit sphere and compared against the fixed antipodal vectors via cosine similarity, yielding layer-wise goodness for weights optimization via directional derivatives.}
% \caption{Schematic of FFzero algorithm, illustrated for a 3-class classification problem. Each linear layer is trained locally and independently by maximizing layer-wise goodness. The cosine similarities of the layer output w.r.t. the fixed prototype vectors (denoted by $\xi[\cdot]$) are used to compute the goodness. Directional derivatives are used to perform gradient-based optimization for each layer locally.}
\label{fig:ff}
\end{figure}

\subsection*{FFzero algorithm}

The core of FFzero is the use of directional-derivative–driven training \cite{baydin2022gradients, ren2022scaling, bandyopadhyay2024single} applied to a modified forward-forward, layer-wise learning strategy \cite{hinton2022forward}. This enables scalable physical learning executed entirely without any form of backpropagation, whether local or global. Integrating the concept of prototype-based learning with large-margin separation\cite{mettes2019hyperspherical}, FFzero generalizes to both classification and regression. Without loss of generality, we continue the exposition using the two-layer neural network example, addressing the classification setting first and then the regression setting.

\subsubsection*{Classification}

We consider the context of multi-class classification with $N$ classes. Let each class label be one-hot encoded into a $N$-dimensional binary vector: $\hat\bfy\in\calY(N)$ with $\calY(N)=\{[1,0,0\dots]^T,[0,1,0\dots]^T,[0,0,1\dots]^T,\dots\}$, wherein the location of `$1$' in the vector identifies the class. Fig.~\ref{fig:ff}(a) shows an illustrative schematic of the classification approach using FFzero.

The key idea of our forward-forward algorithm is to make all the linear module outputs as distinct as possible for each class to enable their identification later. While Hinton \cite{hinton2022forward} and Momeini et al. \cite{momeni2023backpropagation} used positive and negative data with labels appended within the input signal, we introduce a simpler and more general approach that can be scaled to large systems. 

For the first linear layer, we define a set $\xi_1$ of $N$ prototype vectors \cite{mettes2019hyperspherical} on the unit sphere $S^{(\dim(\bfz_1)-1)}$, where $\xi_1[\hat\bfy]$ denotes the prototype associated with the class $\hat\bfy\in\calY(N)$. Each prototype has the same dimensionality as the layer output, i.e.,
$\dim(\xi_1[\cdot]) = \dim(\bfz_1)$, and is defined to be a unit vector, $\|\xi_1[\cdot]\| = 1$. The \textbf{prototypes are chosen randomly, with the constraint that they are maximally separated directionally} (see Supplementary Information 1 for the algorithm of their generation). They are selected prior to training and remain fixed throughout the training process. Analogous to those of the first linear layer, the prototypes $\xi_2$ of the second linear layer (and any subsequent layers, if applicable) can be generated similarly.

A signal tap from each linear layer output is used to compute the \textbf{goodness} $G$ of each class $\hat\bfy$ using the corresponding prototype:
\begin{equation}\label{eq:goodness_classification}
\underbrace{G_1 (\bfW_1|\hat \bfy, \bfx) = \xi_1[\hat \bfy] \cdot \bfz_1/ \|\bfz_1\|}_{\text{class-wise goodness of linear layer \#1}}
\qquad \text{and} 
\qquad
\underbrace{G_2 (\bfW_2|\hat \bfy, \bfx) = \xi_2[\hat \bfy] \cdot \bfz_2/ \|\bfz_2\|}_{\text{class-wise goodness of linear layer \#2}},
\end{equation}
where the goodness is simply the cosine similarity between the layer output and the prototype vector. Given a class label $\hat \bfy$, the goodness function can be geometrically interpreted as how much the output of a linear layer is vectorially (mis-)aligned relative to the corresponding $\xi[\hat \bfy]$. The training objective is to \textit{locally} and \textit{independently} optimize each layer's weights such that the layer output is closest to the prototype of the true label $\bfy_\text{GT}$ and simultaneously farthest away from other prototypes $\calY(N)\setminus\bfy_\text{GT}$. We formalize this layer-wise goodness maximization for classification as:
\be\label{eq:ff_opt}
\begin{aligned}
\bfW_1 & \leftarrow \argmax_{\bfW_1}\ \left( \alpha\ G_1 (\bfW_1| \bfy_\text{GT}, \bfx)\ - \sum_{\hat\bfy\in\calY(N)\setminus \bfy_\text{GT}} G_1 (\bfW_1|\hat \bfy, \bfx)\right) \qquad \text{and} \\
\bfW_2 & \leftarrow \argmax_{\bfW_2}\ \left( \alpha\ G_2 (\bfW_2| \bfy_\text{GT}, \bfx)\ - \sum_{\hat\bfy\in\calY(N)\setminus \bfy_\text{GT}} G_2 (\bfW_2|\hat \bfy, \bfx)\right),
\end{aligned}
\ee
where $\alpha>0$ is a hyperparameter used to weigh the importance of alignment with the correct prototype relative to the other prototypes. We set $\alpha=(N-1)$ in our experiments. In Methods section, we present a slightly modified version of equation \eqref{eq:ff_opt} for additional training stability and efficiency. 

Notably, the optimization in equation \eqref{eq:ff_opt} is local to each layer and decoupled across layers, in contrast to the joint optimization in equation  \eqref{eq:bp-opt} used in classical backpropagation. This local learning allows the nonlinear activation function $\sigma$ to be treated as a complete black box, with no need to know or model its functional form. This is particularly important from a physical standpoint, as nonlinear physical systems, such as optical cavities or mechanical resonators, can be directly employed without constructing surrogate models or computing gradients. Moreover, the physical system need not operate in an element-wise manner, as is common in digital neural networks, and may instead couple multiple dimensions of the input signal.

During inference, each layer votes for the class that attains the maximum class-wise goodness at that layer. For a network with $M$ linear layers, the final network-level prediction is then obtained from the predictions of all layers using a majority voting strategy, 
\be
\bfy \leftarrow \text{mode} \left\{\argmax_{\hat\bfy\in\calY(N)}\ G_m(\bfW_m | \hat\bfy,\bfx): m=1,2,3, \dots, M\right\}.
\ee
The last layer prediction, $\bfy \leftarrow \argmax_{\hat\bfy\in\calY(N)}\ G_M(\bfW_M | \hat\bfy,\bfx)$, is used as a tie-breaking strategy.

For $N$-class classification, inference requires $N$ forward passes through the network. However, because each forward pass can be extremely fast (e.g., operating on picosecond–nanosecond time scales in integrated photonic circuits \cite{shen2017deep, tait2017neuromorphic, miller2017attojoule}) and $N$ is typically small, the resulting increase in inference cost relative to classical single-pass inference is negligible.

\subsubsection*{Regression}
In a regression setting, we first consider the case of a scalar label without loss of generality. Let the label be $\hat y\in[-1,1]$, where min-max normalization has already been performed. Continuing the exposition using the two-layer network, we define a set of two prototype vectors per layer: $\xi_1 = \{\bfxi^l_1,  \bfxi^u_1\}$ and $\xi_2 = \{\bfxi^l_2,  \bfxi^u_2\}$ for the first and second linear layers, respectively. The prototypes are generated similarly to the classification setting. They are random vectors on the sphere $S^{(\dim(\bfz_m)-1)}$ with $\|\bfxi^l_m\|=\|\bfxi^u_m\|=1$ and $\dim(\bfxi^l_m)=\dim(\bfxi^u_m)=\dim(\bfz_m)$ for $m=1,2$. They are also maximally separated from each other, resulting in them being antipodal, i.e., $\bfxi^u_m = -\bfxi^l_m$. The prototypes $\bfxi^l_m$ and $\bfxi^u_m$ are associated with the lower and upper bounds of the label, i.e., $\hat y=-1$ and $\hat y=1$, respectively. Fig.~\ref{fig:ff}(b) illustrates a schematic of the prototype vectors for regression.

Next, we define the goodness based on hyperspherical regression loss. \cite{mettes2019hyperspherical} The ground truth $y_\text{GT}\in[-1,1]$ is interpreted as the set of points on the sphere $S^{(\dim(\bfz_m)-1)}$  whose cosine similarity with $\bfxi^u_m$ equals $y_\text{GT}$. For example, on $S^2$, this corresponds to a circle with an axis along the $\bfxi^u_m$ (see Fig.~\ref{fig:ff}(b)). The training objective is to \textit{locally} and \textit{independently} optimize each layer's weights such that the output of a linear layer is vectorially aligned with the ground truth. Therefore, we define a layer-wise goodness as
\be\label{eq:goodness_regression}
\underbrace{G_1 (\bfW_1|\bfx) = \bfxi_1^u \cdot \bfz_1/ \|\bfz_1\|}_{\text{regression goodness of linear layer \#1}}
\qquad \text{and} 
\qquad
\underbrace{G_2 (\bfW_2|\bfx) = \bfxi_2^u \cdot \bfz_2/ \|\bfz_2\|}_{\text{regression goodness of linear layer \#2}}
\ee
and formalize the regression problem as a layer-wise goodness optimization:
\be\label{eq:ff_opt_regression}
\bfW_1 \leftarrow \argmax_{\bfW_1}\ -\left(G_1 (\bfW_1| \bfx)-y_\text{GT}\right)^2
\qquad \text{and} \qquad
\bfW_2 \leftarrow \argmax_{\bfW_2}\ -\left(G_2 (\bfW_2| \bfx)-y_\text{GT}\right)^2.
\ee
Similar to the classification case, the optimization in equation  \eqref{eq:ff_opt_regression} is local to each layer and remains agnostic to, and unaffected by, the choice of black-box nonlinear activation function. While we discussed the case for regression with a scalar label, it can be extended to multi-dimensional labels by assigning layer-wise prototypes to each label dimension and summing the corresponding goodness functions in equation  \eqref{eq:ff_opt_regression}.

During inference, a single forward pass yields the goodness of each layer, which serves as the prediction of that layer. For a network with $M$ linear layers, the final network-level prediction is then obtained using the last layer prediction, $y \leftarrow G_M(\bfW_M|\bfx)$.

\subsubsection*{Convolutional layers}

For image-based inputs and image recognition tasks, convolutional neural networks (CNNs) are necessary. Consider an expository network architecture as follows:
\begin{equation}
\underbrace{\bfz_1 = \bfW_1\star\bfx}_{\text{convolutional layer \#1}},\qquad \underbrace{\bfu=\sigma(\bfz_1)}_{\text{nonlinear activation}}, \qquad \underbrace{\bfz_2 = \bfW_2 \star \bfu}_{\text{convolutional layer \#2}},
\end{equation}
where $\star$ denotes the convolution operator and $\bfW_1,\bfW_2$ denote trainable convolution kernels. Bias terms are included in practice but omitted here for notational simplicity. Let $\bfx$ be of size $C \times H \times W$ denoting the number of channels, height, and width of the input. $\bfz_1$ is the output of the first convolutional layer with $\text{dim}(\bfz_1) = C_{1} \times H_1 \times W_1$, where $C_{1}$, $H_1$, and $W_1$ are determined by kernel size, stride length, and padding parameters (see Supplementary Information 5.2 for details). $C_1$ also equals the number of kernels in the layer. Each output channel is the result of convolving the input with a distinct learned kernel, followed by bias addition and nonlinearity. $\bfu$ is the result of a nonlinear activation function $\sigma(\cdot)$, subsequently serving as input to the next convolutional layer. The output $\bfz_2$ follows similarly in $\Rset^{C_2\times H_2\times W_2}$.

In contrast to fully connected linear layers, the sizes of $\bfz_1$ and $\bfz_2$ can be significantly larger for convolutional layers, making physical implementation impractical and non-scalable (e.g., requiring extremely high-dimensional prototype vectors in $\Rset^{C_1\times H_1\times W_1}$). To address this issue, we apply dimensionality reduction to $\bfz_1$ prior to the goodness evaluation (see Supplementary Figure 4 for a schematic). Signal taps from $\bfz_1$ and $\bfz_2$ are passed through linear layers with fixed, randomly chosen weights ($\bfA_1$ and $\bfA_2$, respectively) selected prior to training, to reduce dimensionality, i.e.,
\be
\tilde\bfz_1 = \bfA_1 \bfz_1 
\qquad \text{and} \qquad
\tilde\bfz_2 = \bfA_2 \bfz_2,
\ee
where $\dim(\tilde\bfz_1)\ll (C_1\times H_1\times W_1)$ and likewise for the second layer. The prototypes and goodness functions for classification (equation \eqref{eq:goodness_classification}) or regression (equation \eqref{eq:goodness_regression}) are then evaluated on low-dimensional $\tilde\bfz_1$ and $\tilde\bfz_2$, rather than on the original $\bfz_1$ and $\bfz_2$. While dimensionality reduction can be applied to the entire signal, we find that performing dimensionality reduction and the subsequent goodness evaluation independently on each channel yields better performance; this is the approach adopted in our experiments.

\subsubsection*{Generalization to larger architectures}

While the methodology is presented using a two-layer neural network for clarity, FFzero generalizes to deeper architectures, including multilayer perceptrons and convolutional neural networks with arbitrary depth and layer dimensions. The method does not rely on architectural assumptions beyond layer-wise composition. The training and optimization procedure remains unchanged under these architectural generalizations.

\subsubsection*{Training}

Unlike the global optimization (equation  \eqref{eq:bp-opt}) required for backpropagation-based training---where all layers are trained simultaneously---the proposed FFzero framework relies only on local optimization for each linear layer (see equation  \eqref{eq:ff_opt}). However, even these layer-wise optimizations require gradient-based methods for stable training. To address this, we propose using a directional-derivatives-based approach to approximate gradients. 

Directional derivatives are ineffective for global training because gradient approximation errors accumulate across layers, limiting scalability to large neural networks. In contrast, our \textbf{hypothesis} is that for layer-wise local optimizations, the directional-derivative-based approach remains accurate since the errors remain localized and do not propagate between layers.

Let $\max_{\bfomega} \calG(\bfomega)$ denote a generic local goodness-based maximization for any linear or convolutional layer with trainable parameters $\bfomega$, for either classification in equation
\eqref{eq:ff_opt} or regression in equation \eqref{eq:ff_opt_regression}. We propose adapting the iterative training scheme from an exact partial-derivative–based update (computed via automatic differentiation):
\be\label{eq:ff_bp}
\bfomega \leftarrow \bfomega + \lambda \partderiv{\calG(\bfomega)}{\bfomega}
\ee
to a central-difference directional-derivative–based update as:
\be\label{eq:ff_dd}
\begin{aligned}
\bfomega & \leftarrow \bfomega + \lambda\underbrace{n 
\nabla_{\bfv}\calG(\bfomega)
\frac{\bfv}{\|\bfv\|}}_{\text{weights update}}
\qquad \text{with} \quad 
\bfv\sim\calN(\boldsymbol{0},\bfI), 
\quad n = \dim(\bfomega) = \dim(\bfv),
\quad\text{and}\quad
\\
&\text{directional derivative:} \quad \nabla_{\bfv}\calG(\bfomega) \approx \lim_{\varepsilon \rightarrow 0} \frac{\calG(\bfomega + \varepsilon \bfv/\|\bfv\|) - \calG(\bfomega - \varepsilon \bfv/\|\bfv\|)}{2\varepsilon}.
\end{aligned}
\ee

Here, $\bfv$ represents the (unnormalized) direction of simultaneous perturbation for all the weights in $\bfomega$. $\calN(\boldsymbol{0},\bfI)$ denotes the normal distribution with zero mean and identity covariance, which ensures that $\bfv$ is uniformly distributed in direction. The factor $n$ denotes the total number of trainable parameters in $\bfomega$. $\varepsilon$ denotes the magnitude of the perturbation and approaches zero in its limit to ensure accurate approximation. For implementation purposes, the limit can be dropped under the assumption that $\varepsilon\ll 1$. For a schematic illustration, see Fig.~\ref{fig:DD}.

In Supplementary Information 2, we show that the expectation of the directional-derivative update converges to the partial-derivative update, i.e.,
\be\label{eq:dd_convg}
\mathbb{E}_{\bfv}\left[
n 
\nabla_{\bfv}\calG(\bfomega)
\frac{\bfv}{\|\bfv\|}
\right]
= \partderiv{\calG(\bfomega)}{\bfomega}.
\ee
Therefore, even though the directional derivatives may be inaccurate, over many iterations the average optimization trajectory converges to that obtained using partial derivatives (see Fig.~\ref{fig:DD}(c)). In contrast to partial derivatives, which would require one perturbation per trainable parameter (which could be in thousands to millions), directional derivatives only require two forward passes for each direction. Additionally, the factor $n$ (i.e., the number of trainable parameters) in equation  \eqref{eq:ff_dd}  emerges naturally from equation   \eqref{eq:dd_convg} and is critical to maintaining a reasonable learning rate ($\lambda$), especially when $n$ is large. To further improve the accuracy and accelerate this convergence, the directional derivatives may be computed as an average over $P>1$ directions: $\{\bfv_p : p=1,2,\dots,P\}$, i.e.,
\be
\bfomega  \leftarrow \bfomega +\lambda \frac{n}{P} \sum_{p=1}^P
\nabla_{\bfv_p}\calG(\bfomega)
\frac{\bfv_p}{\|\bfv_p\|}.
\label{eq:dd_dir}
\ee
$P$ directions require $2P$ passes. This approach can accelerate convergence and stabilize training. The influence of the number of directions is presented in Supplementary Information 3. Since each forward pass is fast and resource-wise inexpensive, increasing the number of forward passes incurs minimal overhead. We note that while we employ a central difference scheme here, other order schemes are also possible.

\begin{figure}[t]
    \centering
\includegraphics[width=\linewidth]{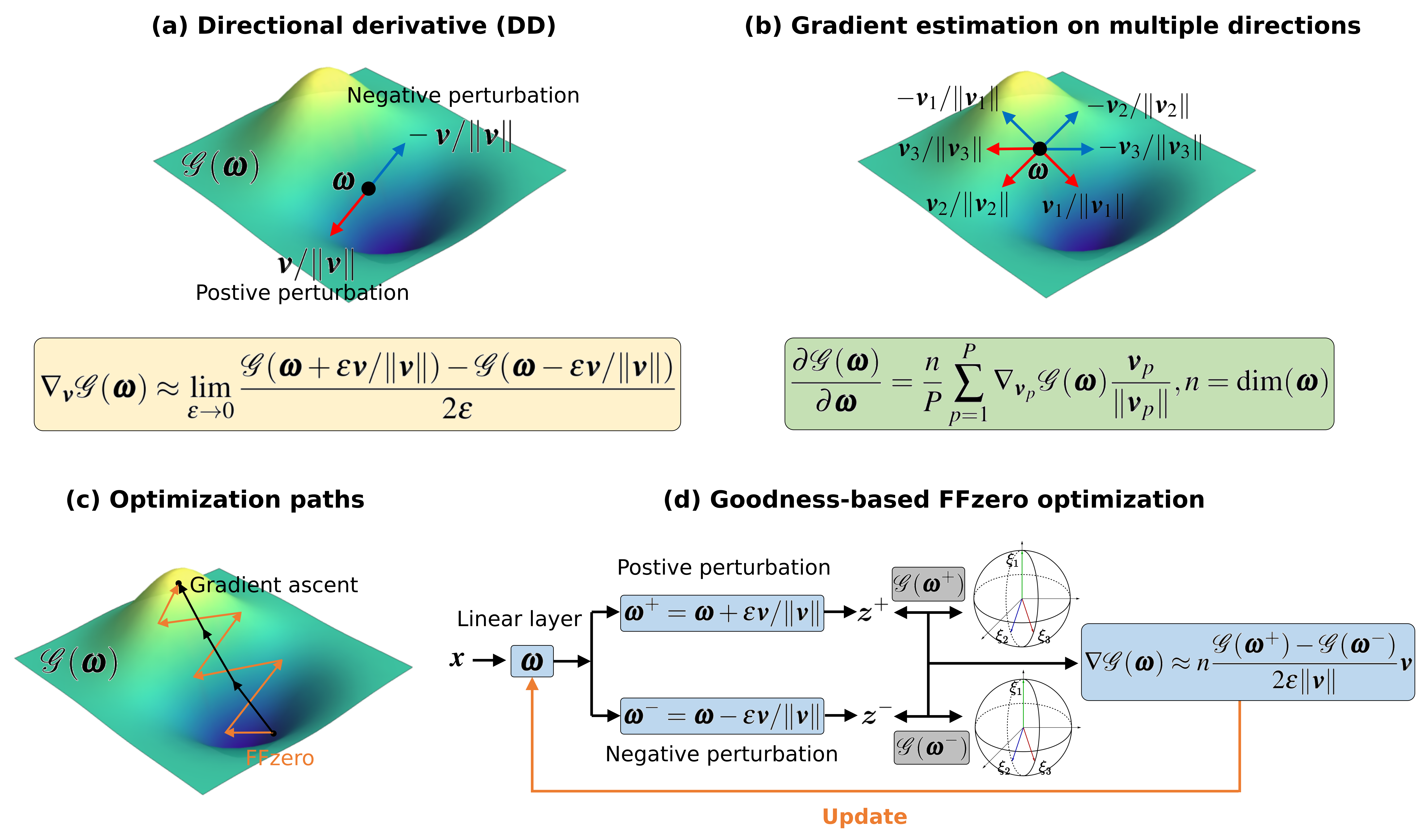}
\caption{Schematic of directional derivative optimization in FFzero. (a) Illustration of directional derivative (DD) estimation on a representative goodness landscape. 
(b) Gradient estimation by averaging directional derivatives over multiple random directions. 
(c) Comparison of optimization trajectories obtained using DD-based updates and standard gradient ascent. While gradient ascent follows the true gradient direction, the expectation of the DD-based update matches the true gradient update (Supplementary Information 2). 
(d) Schematic illustration of the goodness-based DD update for a linear layer.
}
\label{fig:DD}
\end{figure}

\subsection*{Benchmarks}

For comparison, we assess the performance of four different methods, with combinations of backpropagation (BP) vs. forward-forward (FF) training paradigms, and using automatic differentiation (AD) and directional derivatives (DD) as gradient calculation methods. The combinations are BP+AD (the default method for training modern neural networks), BP+DD, FF+AD, and FF+DD, where FF+DD corresponds to our proposed method FFZero. BP+AD serves as an upper-bound performance reference. 

The hyperparameters and experimental settings including learning rate, activation functions, number of directions for calculating directional derivatives ($P$), step size $\varepsilon$, dimension of the prototype vectors, batch size, training epochs, etc. in our experiments are detailed in Methods and Supplementary Information 5. 

\subsubsection*{Performance on image classification with MLPs}

\begin{figure}
    \centering
\includegraphics[width=\linewidth]{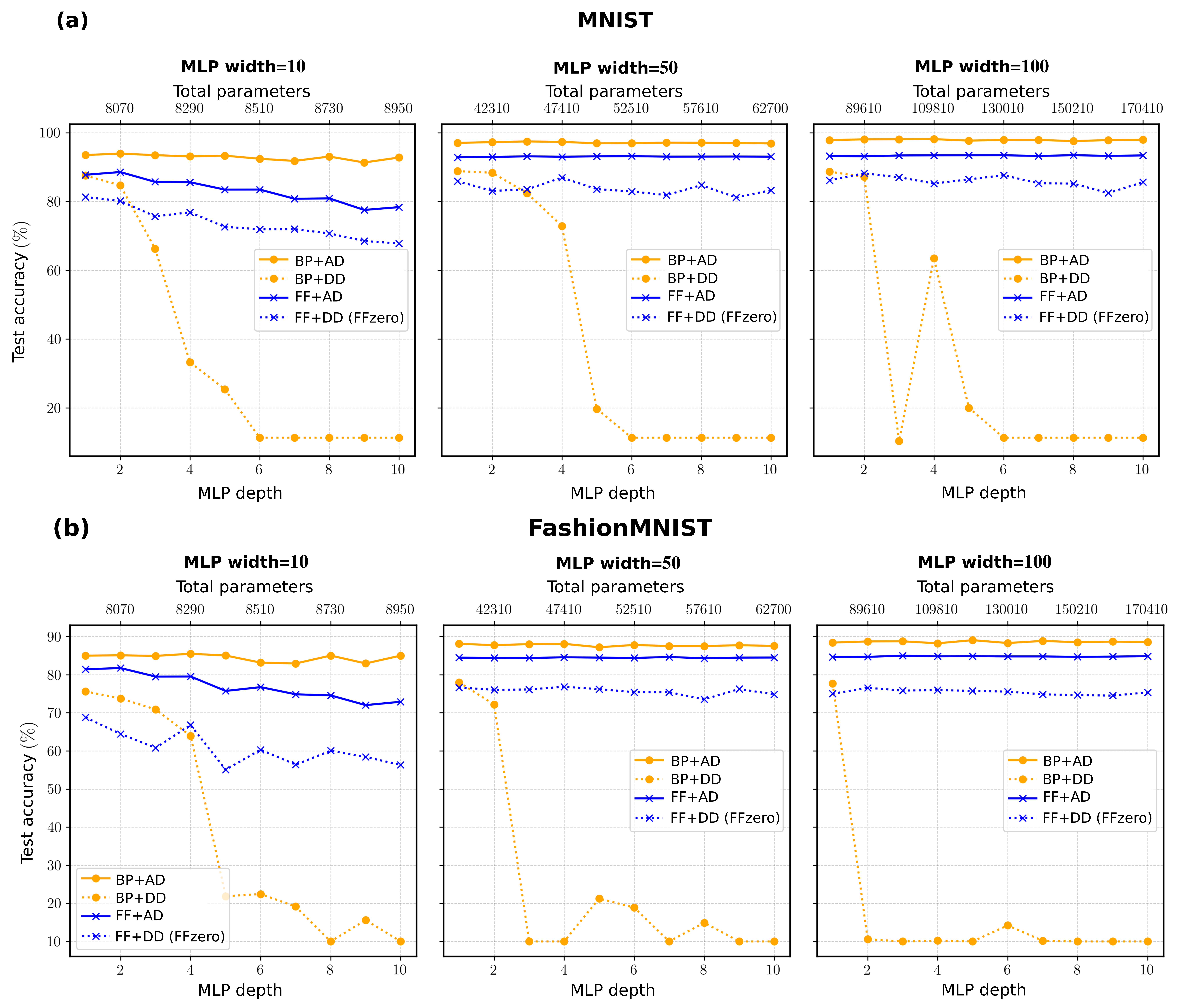}
\caption{Classification accuracy of MLP models across varying hidden-layer widths, with all hidden layers constrained to equal size. Results are shown for (a) MNIST and (b) FashionMNIST datasets using backpropagation (BP) and forward–forward (FF) training paradigms, each combined with either automatic differentiation (AD) or directional-derivative (DD) optimization. The numbers above each plot indicate the total number of parameters for the corresponding model.}
\label{fig:mlp_classification}
\end{figure}

We perform image classification on the MNIST \cite{deng2012mnist} and FashionMNIST \cite{xiao2017fashion} datasets using MLPs of diverse architectures, the results of which are summarized in Fig.~\ref{fig:mlp_classification}.   We show the classification accuracy with respect to network size in terms of the number of layers (depth: 1-10) and the size of each hidden layer (width: 10, 50, 100), using all four combinations of training and gradient calculation methods. 

Across all three network widths, BP+DD performance decreases significantly as the number of layers increases, eventually reaching a level comparable to random guessing, while BP+AD remains stable. This behavior is expected because errors in gradient estimation via directional derivatives accumulate across layers, unlike automatic differentiation that computes exact gradients using computational graphs. In contrast, FF+DD (i.e., our proposed method FFzero) exhibits minimal degradation in accuracy as the network depth increases. This supports the hypothesis that local learning via the forward-forward approach is highly resilient to directional-derivative-based approximations than global backpropagation-based training.

This observation is also reflected in equation   \eqref{eq:dd_dir}, when the number of directions $P$ is fixed, the error in gradient estimate is magnified by the number of trainable parameters $n$. In the layer-wise FF+DD training strategy,
the number of trainable parameters $n$ stays constant as the network becomes deeper. This contrasts with global training via backpropagation, i.e., BP+DD, where the number of trainable parameters increases with depth, resulting in less accurate gradient estimates. 

We note that although both BP+AD and FF+AD outperform FF+DD, the two AD-based methods are physically implausible. Our focus is therefore on DD-based methods, which are physically plausible. The AD-based results are included only for context and as baselines.

\begin{figure}[t]
    \centering
\includegraphics[width=\linewidth]{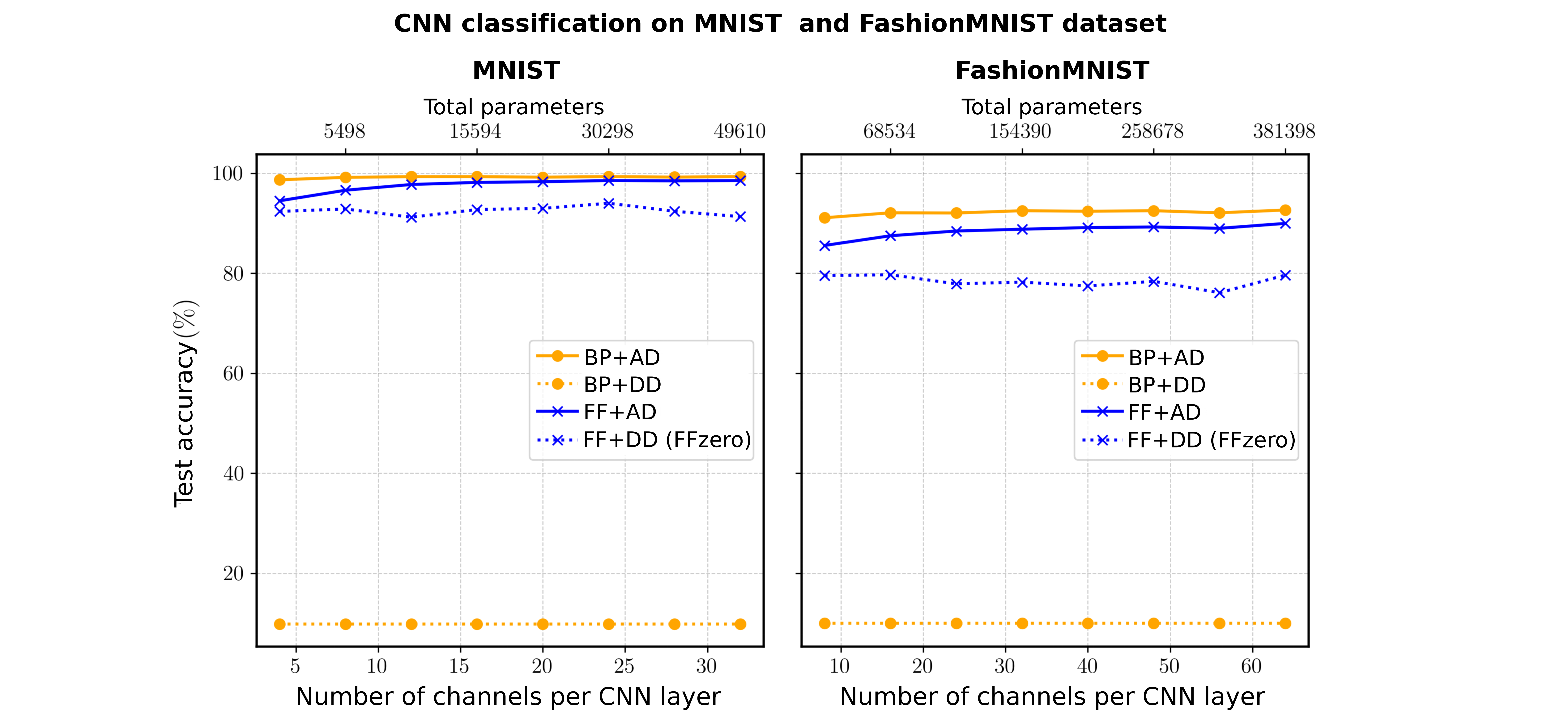}
\caption{Classification accuracy of CNN models across varying number of channels per convolution layer. Results are shown for MNIST and FashionMNIST datasets using backpropagation (BP) and forward–forward (FF) training paradigms, each combined with either automatic differentiation (AD) or directional-derivative (DD) optimization. The numbers above each plot indicate the total number of parameters for the corresponding model.}
\label{fig:cnn-classification}
\end{figure}

\begin{figure}[t]
    \centering
\includegraphics[width=\linewidth]{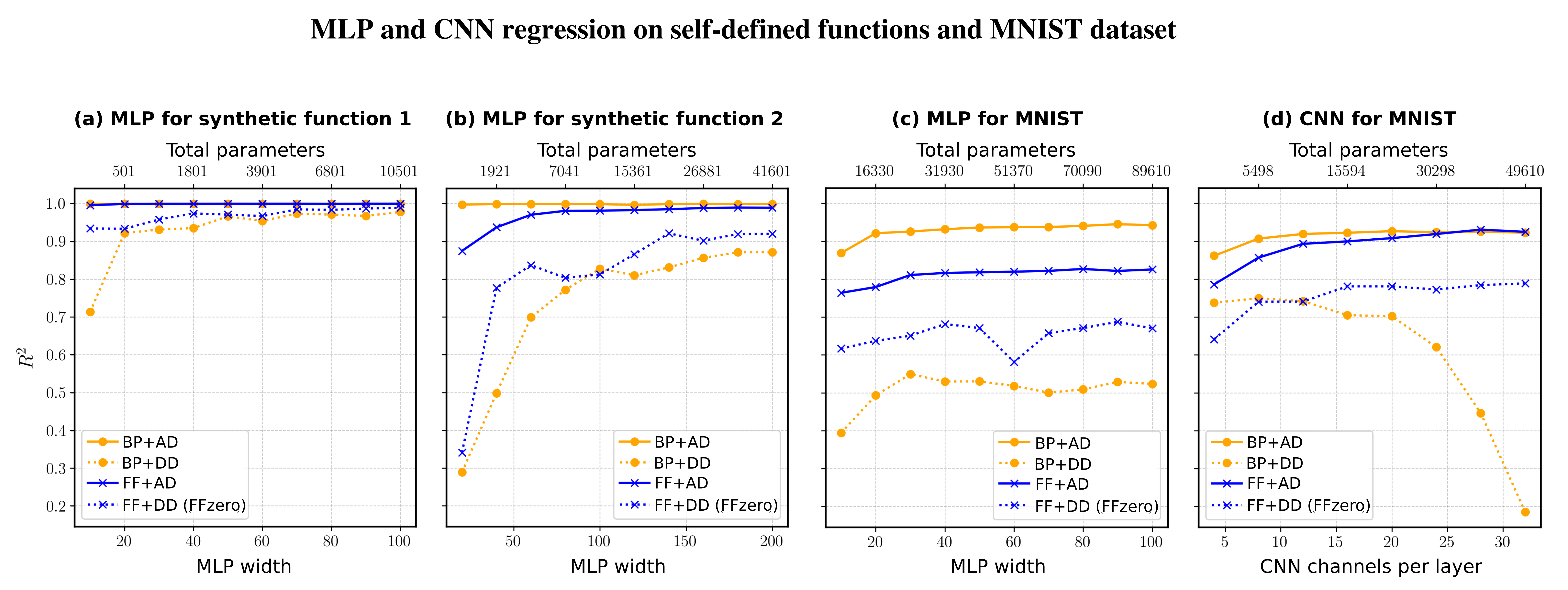}
\caption{Regression accuracy of MLP and CNN models. (a,b) Results are shown for regression on synthetic functions in equation  \eqref{eq:regression_function} using MLPs across varying hidden-layer widths, with all hidden layers constrained to equal size. (c,d) Results are shown for regression on MNIST using MLP and CNN of varying hidden-layer widths and number of channels per convolution layer, respectively. For all cases, the comparison is made between backpropagation (BP) and forward–forward (FF) training paradigms, each combined with either automatic differentiation (AD) or directional-derivative (DD) optimization. The numbers above each plot indicate the total number of parameters for the corresponding model.} 
\label{fig:regression}
\end{figure}

\subsubsection*{Performance on image classification with CNNs} 

Fig.~\ref{fig:cnn-classification} shows classification results on the MNIST and FashionMNIST datasets using CNN architecture with two convolutional layers (followed by a fully-connected linear layer) and varying number of channels per layer. With channel-wise training, the accuracy of FF+DD shows a trend similar to that observed with the MLP architecture. As the number of CNN channels per layer increases (resulting in increasing number of parameters), the performance of FF+DD remains stable. In contrast, BP+DD is extremely sensitive to the errors from directional derivatives. In CNN, even with a smaller number of parameters, the models suffer from erroneous gradient estimations more than in the MLP case, due to the architectural design that extracts local structure via the convolutional kernels. This leads to models trained with BP+DD exhibiting near-chance performance.

\subsubsection*{Performance on function and image regression} 

Fig.~\ref{fig:regression} summarizes the model performance on a series of regression tasks across multiple datasets. Regression performance is quantified using the coefficient of determination $R^2$. We design two function regression benchmarks with 2D and 5D inputs corresponding to  increasing input dimensionality: 
\be\label{eq:regression_function}
\begin{aligned}
& \text{ synthetic function 1: } y(x_1, x_2) = \sin(x_1) + \cos(x_2) \text{ and } \\
& \text{ synthetic function 2: } y(x_1,x_2,x_3, x_4, x_5) = e^{x_1} \sin (x_2) + x_3 \cos(x_4) - x_5x_1, 
\end{aligned}
\ee
where $y$ denotes the function value and $x_{(\cdot)}$ represents the input values. Fig.~\ref{fig:regression}(a)-(b) demonstrates the function regression accuracy for FFzero using a MLP with hidden-layer depth of 2 and varying hidden-layer width. Further, we benchmark FFZero on  MNIST formulated as a regression task in Fig.~\ref{fig:regression}(c)-(d) using a MLP and CNN, respectively. Concretely, each digit image corresponds to a real-valued label between 0 and 9, and the model outputs a single scalar prediction. Across both synthetic and MNIST regression settings, FF+DD consistently outperforms BP+DD as the number of network parameters increases, indicating more stable scaling behavior with respect to model size.

\subsection*{Example of physical implementation of FFzero}

To demonstrate the physical implementability of FFzero, we emulate a photonic neural network in silico. For this purpose, we use the Neurophox framework,\cite{pai2019matrix, williamson2019reprogrammable} an open-source simulation library for programmable photonic networks.

\begin{figure}
    \centering
\includegraphics[width=\linewidth]{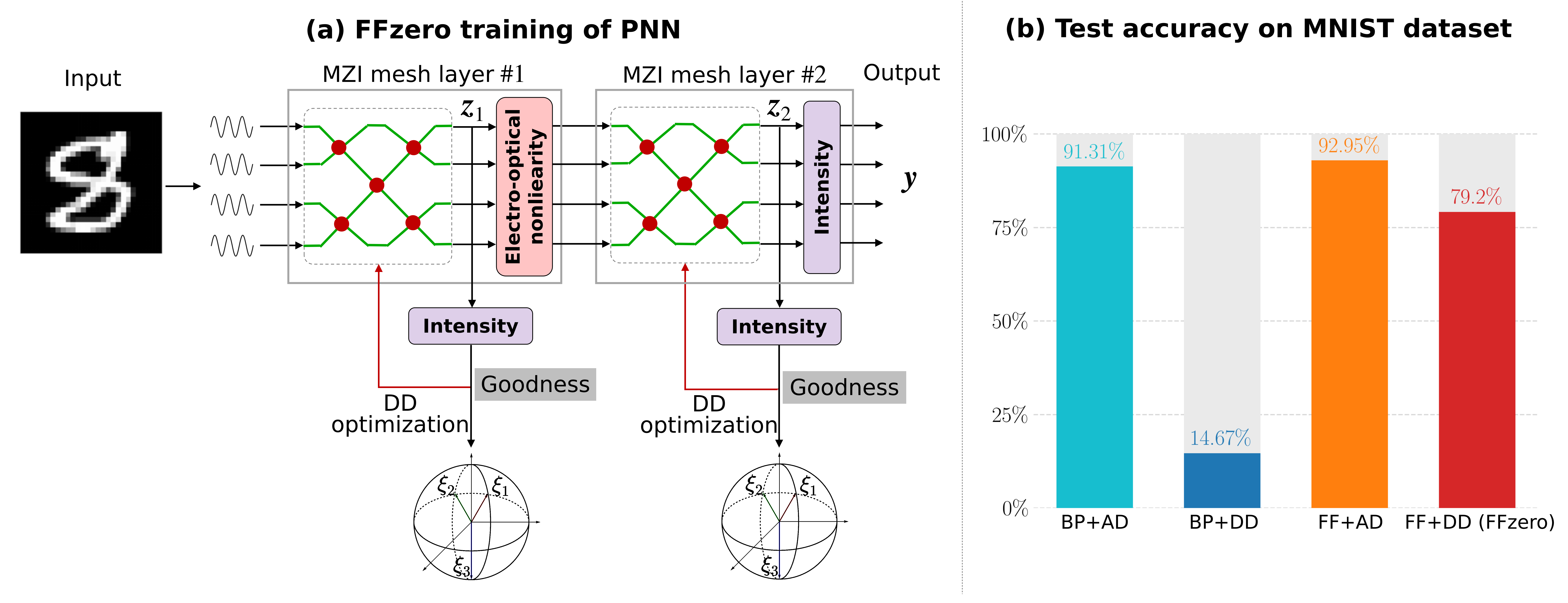}
\caption{
\textbf{(a)}~Schematic of FFzero training applied to a two-layer (emulated) photonic
neural network. 
\textbf{(b)}~Test accuracy on MNIST using BP+AD, BP+DD, FF+AD, and FF+DD (FFzero).
}
\label{fig:PNN}
\end{figure}

We model a two-layer photonic neural network. A linear layer consists of a programmable rectangular Mach-Zehnder interferometer (MZI) mesh with Clements decomposition\cite{clements2016optimal} to perform matrix-vector multiplication with trainable weights. Each MZI unit cell in the mesh is parameterized by phase shifts $\theta \in [0, \pi]$, $\varphi \in [0, 2\pi)$, and  $\gamma \in [0, 2\pi)$ representing a unitary transformation. An electro-optical nonlinearity\cite{williamson2019reprogrammable} is interleaved between the linear layers. We refer to the network of ref.\cite{pai2019parallelfaulttolerantprogrammingarbitrary,pai2019matrix,williamson2019reprogrammable} for details of the MZI and electro-optical nonlinearity implementation and show a schematic of the network in Fig.~\ref{fig:PNN}(a).

We apply the FFzero framework to train the photonic linear layers for image classification following the layer-wise local learning strategy described previously.  Within FFzero, the gradients are not propagated through the electro-optical nonlinearity---consistent with the FFzero principle that nonlinearities are treated as unknown physical transformations whose internal dynamics need not be modelled.

At the network input, the grayscale pixel values of the image are mapped to the amplitudes of the input signals, with all signals being in phase. For each linear layer, the intensity at each output port is used in the goodness computation. The goodness is calculated based on pre-computed prototype vectors as discussed previously. The phase parameters $\{\theta, \varphi, \gamma\}$ for each MZI unit cells in each linear layer are optimized to maximize the goodness using the cosine similarity-based loss defined in equation \eqref{eq:ff_opt}.

We benchmark this emulation of FFzero on MNIST, where the input is a $28\times28$ grayscale image. The two linear layers therefore represent transformations in a 
$784\times784$ dimensional space. For benchmarking purposes, gradient tracking in the emulator can be used to enable automatic differentiation, which can be compared against directional-derivatives-based approaches. As baselines, we perform backpropagation experiments (BP+AD and BP+DD), where the intensities at the first 10 output ports are selected and a cross-entropy loss is applied directly. For FF+AD and FF+DD (i.e., FFzero), all output ports are retained for the prototype-based goodness computation.

Fig.~\ref{fig:PNN}(b) summarizes classification accuracy on the MNIST test dataset. As expected, FF+DD (FFzero) outperforms BP+DD significantly, confirming that combining
directional-derivatives-based optimization with layer-wise local learning mitigates the gradient-estimation errors that cause BP+DD to collapse, even in photonics emulation.  Additional analysis of the layer embeddings extracted from the output of each MZI mesh in FFzero (Supplementary Information 4) indicates that the embeddings cluster around the prototype vectors of the corresponding classes.

These findings  support that FFzero can be adapted on contemporary photonic neural network architectures (e.g., those using MZIs\cite{Ashtiani2022,bandyopadhyay2024single})  with minimal modifications. The framework requires no knowledge of the nonlinearity during training, relying solely on forward evaluations of the network. This opens the possibility for true in-situ training of programmable photonic chips. Although demonstrated for photonics, FFzero is a general-purpose, backpropagation-free learning framework applicable to diverse physical modalities, including mechanical, electrical, and acoustic systems.

\section*{Discussion}

With the growing interest in alternative computing paradigms such as physical and neuromorphic computing, moving beyond backpropagation-based training remains an important objective. In this light, FFzero introduces a general-purpose, backpropagation-free algorithm for training neural networks. The FFzero approach opens possibilities for physical neural network implementations capable of performing both training and inference directly in hardware, unlike many existing approaches that rely on digital twins or auxiliary computations executed on conventional CPUs or GPUs.

Under the premise that automatic differentiation is physically implausible in many physical systems, FFzero employs directional derivatives as a zeroth-order approximation to the gradient for gradient-ascent-based training. At the same time, it halts the global error propagation of directional derivatives through chain rule by adopting a local, layer-by-layer training strategy. This results in a training procedure where weight updates are driven by local, forward-only passes that are physically implementable and computationally efficient, while remaining agnostic to the choice and knowledge of nonlinearities. This property is particularly attractive for physical systems such as photonic neural networks, where a forward pass can be executed in hundreds of picoseconds, \cite{Ashtiani2022} enabling billions of forward passes per second with minimal energy consumption and latency.

From an algorithmic perspective, FFzero is  generalizable to different network architectures, supporting both classification and regression tasks, and accommodating diverse input modalities. It is also feasible to explore extensions that increase the number of classes or output dimensions without requiring relearning of existing prototype alignments, potentially enabling incremental or continual learning scenarios. Looking ahead, FFzero could potentially be adapted to train transformer-based architectures and generative models, offering a pathway to mitigate their current speed and energy bottlenecks.

Although FFzero does not yet match the optimization performance and accuracy of backpropagation-based methods---since directional derivative estimation and local weight updates are inherently less precise than automatic differentiation and global backpropagation---it offers a unique trade-off. Specifically, FFzero sacrifices some accuracy in exchange for significantly improved physical implementability. Further experimental validation through hardware realizations therefore represents an important next step. Such implementations, among many possible physical systems, could build upon existing photonic neural network platforms,\cite{bandyopadhyay2024single,Ashtiani2022} which employ Mach-Zehnder interferometer meshes for linear transformations and photonic resonators for nonlinear activation functions. Another promising direction is the adaptation to iontronic networks\cite{Conte2025} based on microfluidic ionic channels, which could enable in-situ learning in bionic applications.  These extensions underscore the potential of FFzero for stable backpropagation-free physical learning.

\section*{Methods}

\subsubsection*{Classification loss}

For classification tasks, the parameters of the $m^\text{th}$ layer in an MLP, or equivalently the parameters of $m^\text{th}$ channel in a CNN layer, are optimized using a margin-based loss. The optimization problem is formulated as
\be
\label{eq:loss_classification}
\bfW_{m} \leftarrow
\argmin_{\bfW_{m}}
\sum_{\hat{\bfy} \in \mathcal{Y}(N)\setminus \bfy_{\mathrm{GT}}}
\mathrm{ReLU}\Bigl(
G_{m}(\bfW_{m} \mid \hat{\bfy}, \bfx)
- G_{m}(\bfW_{m} \mid \bfy_{\mathrm{GT}}, \bfx)
+ q
\Bigr),
\ee
where $q$ denotes the margin parameter. The margin value and the use of the ReLU activation are chosen empirically to improve training stability and enhance discriminative feature learning; both can be adjusted depending on the specific application.

\subsubsection*{FFzero training strategy}

In the FFzero framework, network layers are trained sequentially and independently. When optimizing a given layer, all previously trained layers are kept fixed and are not updated. For a fair comparison with backpropagation (BP), each layer in the forward–forward (FF) setting is trained for the same number of epochs as in BP.

\subsubsection*{Implementation and benchmark details}

Supplementary Information 5 provides details on the implementation and benchmarking of FFzero and backpropagation methods. All parameters and hyperparameters used in the experiments are listed in Supplementary Table 1.

\section*{Code availability}
The codes generated in this study are available at: \href{https://github.com/mmc-group/FFzero-local-learning}{https://github.com/mmc-group/FFzero-local-learning}.

\section*{Data availability}
Only publicly available datasets were used, and no dataset was generated in this study.

% \section*{Author contributions}

\section*{Competing interests}
The authors declare no competing interests.

\bibliography{Bib}

@article{deng2012mnist,
  title={The mnist database of handwritten digit images for machine learning research [best of the web]},
  author={Deng, Li},
  journal={IEEE signal processing magazine},
  volume={29},
  number={6},
  pages={141--142},
  year={2012},
  publisher={IEEE}
}

@article{xiao2017fashion,
  title={Fashion-mnist: a novel image dataset for benchmarking machine learning algorithms},
  author={Xiao, Han and Rasul, Kashif and Vollgraf, Roland},
  journal={Preprint at:  	
  https://doi.org/10.48550/arXiv.1708.07747},
  year={2017},
}

@article{chen2025much,
  title={How much energy will AI really consume? The good, the bad and the unknown},
  author={Chen, Sophia},
  journal={Nature},
  volume={639},
  number={8053},
  pages={22--24},
  year={2025},
  publisher={Nature}
}

@inproceedings{strubell2019energy,
  title={Energy and policy considerations for deep learning in NLP},
  author={Strubell, Emma and Ganesh, Ananya and McCallum, Andrew},
  booktitle={Proceedings of the 57th annual meeting of the association for computational linguistics},
  pages={3645--3650},
  year={2019}
}

@inproceedings{gowda2024watt,
  title={Watt for what: Rethinking deep learning’s energy-performance relationship},
  author={Gowda, Shreyank N and Hao, Xinyue and Li, Gen and Gowda, Shashank Narayana and Jin, Xiaobo and Sevilla-Lara, Laura},
  booktitle={European Conference on Computer Vision},
  pages={388--405},
  year={2024},
  organization={Springer}
}

@article{ye2025overhead,
  title={An overhead-reduced, efficient, fully analog neural-network computing hardware},
  author={Ye, Jiabao and Wang, Wannian and Shi, Caiping and Cui, Xuecheng and Qu, Wanyuan and Lin, Peng and Yu, Xiao and Cheng, Ran and Wu, Hanming and Chen, Bing},
  journal={Science Advances},
  volume={11},
  number={43},
  pages={eadv7555},
  year={2025},
  publisher={American Association for the Advancement of Science}
}

@article{tanaka2019recent,
  title={Recent advances in physical reservoir computing: A review},
  author={Tanaka, Gouhei and Yamane, Toshiyuki and H{\'e}roux, Jean Benoit and Nakane, Ryosho and Kanazawa, Naoki and Takeda, Seiji and Numata, Hidetoshi and Nakano, Daiju and Hirose, Akira},
  journal={Neural Networks},
  volume={115},
  pages={100--123},
  year={2019},
  publisher={Elsevier}
}

@article{kiyabu2025optomechanical,
  title={Optomechanical reservoir computing},
  author={Kiyabu, Steven and Nelson, Daniel and Thomson, John and Schultz, Benjamin and Vincent, Timothy and Hertlein, Nathan and Gillman, Andrew and Criner, Amanda and Buskohl, Philip R},
  journal={Proceedings of the National Academy of Sciences},
  volume={122},
  number={29},
  pages={e2424991122},
  year={2025},
  publisher={National Academy of Sciences}
}

@inproceedings{nelson2024spectral,
  title={Spectral Analysis of Mechanical Reservoir Computing With ReLU Spring Networks},
  author={Nelson, Daniel and Kiyabu, Steven and Vincent, Timothy and Gillman, Andrew and Criner, Amanda and Buskohl, Philip R},
  booktitle={Smart Materials, Adaptive Structures and Intelligent Systems},
  volume={88322},
  pages={V001T08A004},
  year={2024},
  organization={American Society of Mechanical Engineers}
}

@article{van2017advances,
  title={Advances in photonic reservoir computing},
  author={Van der Sande, Guy and Brunner, Daniel and Soriano, Miguel C},
  journal={Nanophotonics},
  volume={6},
  number={3},
  pages={561--576},
  year={2017},
  publisher={De Gruyter},
}

@article{guo2024mems,
  title={MEMS reservoir computing system with stiffness modulation for multi-scene data processing at the edge},
  author={Guo, Xiaowei and Yang, Wuhao and Xiong, Xingyin and Wang, Zheng and Zou, Xudong},
  journal={Microsystems \& Nanoengineering},
  volume={10},
  number={1},
  pages={84},
  year={2024},
  publisher={Nature Publishing Group UK London}
}

@incollection{taniguchi2021reservoir,
  title={Reservoir computing based on spintronics technology},
  author={Taniguchi, Tomohiro and Tsunegi, Sumito and Miwa, Shinji and Fujii, Keisuke and Kubota, Hitoshi and Nakajima, Kohei},
  booktitle={Reservoir computing: theory, physical implementations, and applications},
  pages={331--360},
  year={2021},
  publisher={Springer}
}

@article{appeltant2011information,
  title={Information processing using a single dynamical node as complex system},
  author={Appeltant, Lennert and Soriano, Miguel Cornelles and Van der Sande, Guy and Danckaert, Jan and Massar, Serge and Dambre, Joni and Schrauwen, Benjamin and Mirasso, Claudio R and Fischer, Ingo},
  journal={Nature communications},
  volume={2},
  number={1},
  pages={468},
  year={2011},
  publisher={Nature Publishing Group UK London}
}

@article{yan2024emerging,
  title={Emerging opportunities and challenges for the future of reservoir computing},
  author={Yan, Min and Huang, Can and Bienstman, Peter and Tino, Peter and Lin, Wei and Sun, Jie},
  journal={Nature Communications},
  volume={15},
  number={1},
  pages={2056},
  year={2024},
  publisher={Nature Publishing Group UK London}
}

@inproceedings{Kasarla2022,
 author = {Kasarla, Tejaswi and Burghouts, Gertjan and van Spengler, Max and van der Pol, Elise and Cucchiara, Rita and Mettes, Pascal},
 booktitle = {Advances in Neural Information Processing Systems},
 editor = {S. Koyejo and S. Mohamed and A. Agarwal and D. Belgrave and K. Cho and A. Oh},
 pages = {19553--19566},
 publisher = {Curran Associates, Inc.},
 title = {Maximum Class Separation as Inductive Bias in One Matrix},
 url = {https://proceedings.neurips.cc/paper_files/paper/2022/file/7b95e1ca9d7347da59cefd362e60a0b6-Paper-Conference.pdf},
 volume = {35},
 year = {2022}
}

@article{mezzadri2006,
  title={How to generate random matrices from the classical compact groups},
  author={Mezzadri, Francesco},
  journal={Preprint at:  	
  https://doi.org/10.48550/arXiv.math-ph/0609050},
  year={2006}
}

@article{zidan2018future,
  title={The future of electronics based on memristive systems},
  author={Zidan, Mohammed A and Strachan, John Paul and Lu, Wei D},
  journal={Nature electronics},
  volume={1},
  number={1},
  pages={22--29},
  year={2018},
  publisher={Nature Publishing Group UK London}
}

@inproceedings{ganguly2019towards,
  title={Towards energy efficient non-von neumann architectures for deep learning},
  author={Ganguly, Antara and Muralidhar, Rajeev and Singh, Virendra},
  booktitle={20th international symposium on quality electronic design (ISQED)},
  pages={335--342},
  year={2019},
  organization={IEEE}
}

@article{heller1978survey,
  title={A survey of parallel algorithms in numerical linear algebra},
  author={Heller, Don},
  journal={Siam Review},
  volume={20},
  number={4},
  pages={740--777},
  year={1978},
  publisher={SIAM}
}

@article{kimovski2023beyond,
  title={Beyond von neumann in the computing continuum: Architectures, applications, and future directions},
  author={Kimovski, Dragi and Saurabh, Nishant and Jansen, Matthijs and Aral, Atakan and Al-Dulaimy, Auday and Bondi, Andr{\'e} B and Galletta, Antonino and Papadopoulos, Alessandro V and Iosup, Alexandru and Prodan, Radu},
  journal={IEEE Internet Computing},
  volume={28},
  number={3},
  pages={6--16},
  year={2023},
  publisher={IEEE}
}

@article{seok2024beyond,
  title={Beyond von Neumann architecture: Brain-inspired artificial neuromorphic devices and integrated computing},
  author={Seok, Hyunho and Lee, Dongho and Son, Sihoon and Choi, Hyunbin and Kim, Gunhyoung and Kim, Taesung},
  journal={Advanced Electronic Materials},
  volume={10},
  number={8},
  pages={2300839},
  year={2024},
  publisher={Wiley Online Library}
}

@article{mondal2024recent,
  title={Recent trends in neuromorphic systems for non-von Neumann in materia computing and cognitive functionalities},
  author={Mondal, Indrajit and Attri, Rohit and Rao, Tejaswini S and Yadav, Bhupesh and Kulkarni, Giridhar U},
  journal={Applied Physics Reviews},
  volume={11},
  number={4},
  year={2024},
  publisher={AIP Publishing}
}

@article{momeni2025training,
  title={Training of physical neural networks},
  author={Momeni, Ali and Rahmani, Babak and Scellier, Benjamin and Wright, Logan G and McMahon, Peter L and Wanjura, Clara C and Li, Yuhang and Skalli, Anas and Berloff, Natalia G and Onodera, Tatsuhiro and others},
  journal={Nature},
  volume={645},
  number={8079},
  pages={53--61},
  year={2025},
  publisher={Nature Publishing Group UK London}
}

@article{wright2022deep,
  title={Deep physical neural networks trained with backpropagation},
  author={Wright, Logan G and Onodera, Tatsuhiro and Stein, Martin M and Wang, Tianyu and Schachter, Darren T and Hu, Zoey and McMahon, Peter L},
  journal={Nature},
  volume={601},
  number={7894},
  pages={549--555},
  year={2022},
  publisher={Nature Publishing Group UK London}
}

@article{nakajima2022physical,
  title={Physical deep learning with biologically inspired training method: gradient-free approach for physical hardware},
  author={Nakajima, Mitsumasa and Inoue, Katsuma and Tanaka, Kenji and Kuniyoshi, Yasuo and Hashimoto, Toshikazu and Nakajima, Kohei},
  journal={Nature communications},
  volume={13},
  number={1},
  pages={7847},
  year={2022},
  publisher={Nature Publishing Group UK London}
}

@article{bandyopadhyay2024single,
  title={Single-chip photonic deep neural network with forward-only training},
  author={Bandyopadhyay, Saumil and Sludds, Alexander and Krastanov, Stefan and Hamerly, Ryan and Harris, Nicholas and Bunandar, Darius and Streshinsky, Matthew and Hochberg, Michael and Englund, Dirk},
  journal={Nature Photonics},
  volume={18},
  number={12},
  pages={1335--1343},
  year={2024},
  publisher={Nature Publishing Group UK London}
}

@article{spall2002implementation,
  title={Implementation of the simultaneous perturbation algorithm for stochastic optimization},
  author={Spall, James C},
  journal={IEEE Transactions on aerospace and electronic systems},
  volume={34},
  number={3},
  pages={817--823},
  year={2002},
  publisher={IEEE}
}

@article{yin2001random,
  title={Random-direction optimization algorithms with applications to threshold controls},
  author={Yin, G and Zhang, Q and Yan, HM and Boukas, El-K{\'e}bir},
  journal={Journal of optimization theory and applications},
  volume={110},
  number={1},
  pages={211--233},
  year={2001},
  publisher={Springer}
}

@article{larson2019derivative,
  title={Derivative-free optimization methods},
  author={Larson, Jeffrey and Menickelly, Matt and Wild, Stefan M},
  journal={Acta Numerica},
  volume={28},
  pages={287--404},
  year={2019},
  publisher={Cambridge University Press}
}

@article{hinton2022forward,
  title={The forward-forward algorithm: Some preliminary investigations},
  author={Hinton, Geoffrey},
  journal={Preprint at:  	
  https://doi.org/10.48550/arXiv.2212.13345},
  year={2022}
}

@article{nokland2016direct,
  title={Direct feedback alignment provides learning in deep neural networks},
  author={N{\o}kland, Arild},
  journal={Advances in neural information processing systems},
  volume={29},
  year={2016}
}

@article{momeni2023backpropagation,
  title={Backpropagation-free training of deep physical neural networks},
  author={Momeni, Ali and Rahmani, Babak and Mall{\'e}jac, Matthieu and Del Hougne, Philipp and Fleury, Romain},
  journal={Science},
  volume={382},
  number={6676},
  pages={1297--1303},
  year={2023},
  publisher={American Association for the Advancement of Science}
}

@article{laydevant2024training,
  title={Training an ising machine with equilibrium propagation},
  author={Laydevant, J{\'e}r{\'e}mie and Markovi{\'c}, Danijela and Grollier, Julie},
  journal={Nature Communications},
  volume={15},
  number={1},
  pages={3671},
  year={2024},
  publisher={Nature Publishing Group UK London}
}

@article{journe2022hebbian,
  title={Hebbian deep learning without feedback},
  author={Journ{\'e}, Adrien and Rodriguez, Hector Garcia and Guo, Qinghai and Moraitis, Timoleon},
  journal={Preprint at: 	
  https://doi.org/10.48550/arXiv.2209.11883},
  year={2022}
}

@article{baydin2022gradients,
  title={Gradients without backpropagation},
  author={Baydin, At{\i}l{\i}m G{\"u}ne{\c{s}} and Pearlmutter, Barak A and Syme, Don and Wood, Frank and Torr, Philip},
  journal={Preprint at:  	
  https://doi.org/10.48550/arXiv.2202.08587},
  year={2022}
}

@inproceedings{paszke2017automatic,
  title={Automatic differentiation in PyTorch},
  author={Paszke, Adam and Gross, Sam and Chintala, Soumith and Chanan, Gregory and Yang, Edward and DeVito, Zachary and Lin, Zeming and Desmaison, Alban and Antiga, Luca and Lerer, Adam},
  booktitle={NIPS-W},
  year={2017}
}

@article{shalf2015computing,
  title={Computing beyond moore's law},
  author={Shalf, John M and Leland, Robert},
  journal={Computer},
  volume={48},
  number={12},
  pages={14--23},
  year={2015},
  publisher={IEEE}
}

@article{nakajima2020physical,
  title={Physical reservoir computing—an introductory perspective},
  author={Nakajima, Kohei},
  journal={Japanese Journal of Applied Physics},
  volume={59},
  number={6},
  pages={060501},
  year={2020},
  publisher={IOP Publishing}
}

@article{nakajima2021scalable,
  title={Scalable reservoir computing on coherent linear photonic processor},
  author={Nakajima, Mitsumasa and Tanaka, Kenji and Hashimoto, Toshikazu},
  journal={Communications Physics},
  volume={4},
  number={1},
  pages={20},
  year={2021},
  publisher={Nature Publishing Group UK London}
}

@article{moon2021hierarchical,
  title={Hierarchical architectures in reservoir computing systems},
  author={Moon, John and Wu, Yuting and Lu, Wei D},
  journal={Neuromorphic Computing and Engineering},
  volume={1},
  number={1},
  pages={014006},
  year={2021},
  publisher={IOP Publishing}
}

@article{rumelhart1986learning,
  title={Learning representations by back-propagating errors},
  author={Rumelhart, David E and Hinton, Geoffrey E and Williams, Ronald J},
  journal={nature},
  volume={323},
  number={6088},
  pages={533--536},
  year={1986},
  publisher={Nature Publishing Group UK London}
}

@article{lillicrap2020backpropagation,
  title={Backpropagation and the brain},
  author={Lillicrap, Timothy P and Santoro, Adam and Marris, Luke and Akerman, Colin J and Hinton, Geoffrey},
  journal={Nature Reviews Neuroscience},
  volume={21},
  number={6},
  pages={335--346},
  year={2020},
  publisher={Nature Publishing Group UK London}
}

@article{ambrogio2018equivalent,
  title={Equivalent-accuracy accelerated neural-network training using analogue memory},
  author={Ambrogio, Stefano and Narayanan, Pritish and Tsai, Hsinyu and Shelby, Robert M and Boybat, Irem and Di Nolfo, Carmelo and Sidler, Severin and Giordano, Massimo and Bodini, Martina and Farinha, Nathan CP and others},
  journal={Nature},
  volume={558},
  number={7708},
  pages={60--67},
  year={2018},
  publisher={Nature Publishing Group UK London}
}

@article{laborieux2022holomorphic,
  title={Holomorphic equilibrium propagation computes exact gradients through finite size oscillations},
  author={Laborieux, Axel and Zenke, Friedemann},
  journal={Advances in neural information processing systems},
  volume={35},
  pages={12950--12963},
  year={2022}
}

@article{mettes2019hyperspherical,
  title={Hyperspherical prototype networks},
  author={Mettes, Pascal and Van der Pol, Elise and Snoek, Cees},
  journal={Advances in neural information processing systems},
  volume={32},
  year={2019}
}

@article{ren2022scaling,
  title={Scaling forward gradient with local losses},
  author={Ren, Mengye and Kornblith, Simon and Liao, Renjie and Hinton, Geoffrey},
  journal={Preprint at:  	
  https://doi.org/10.48550/arXiv.2210.03310},
  year={2022}
}

@article{shen2017deep,
  title={Deep learning with coherent nanophotonic circuits},
  author={Shen, Yichen and Harris, Nicholas C and Skirlo, Scott and Prabhu, Mihika and Baehr-Jones, Tom and Hochberg, Michael and Sun, Xin and Zhao, Shijie and Larochelle, Hugo and Englund, Dirk and others},
  journal={Nature photonics},
  volume={11},
  number={7},
  pages={441--446},
  year={2017},
  publisher={Nature Publishing Group UK London}
}

@article{tait2017neuromorphic,
  title={Neuromorphic photonic networks using silicon photonic weight banks},
  author={Tait, Alexander N and De Lima, Thomas Ferreira and Zhou, Ellen and Wu, Allie X and Nahmias, Mitchell A and Shastri, Bhavin J and Prucnal, Paul R},
  journal={Scientific reports},
  volume={7},
  number={1},
  pages={7430},
  year={2017},
  publisher={Nature Publishing Group UK London}
}

@article{miller2017attojoule,
  title={Attojoule optoelectronics for low-energy information processing and communications},
  author={Miller, David AB},
  journal={Journal of Lightwave Technology},
  volume={35},
  number={3},
  pages={346--396},
  year={2017},
  publisher={IEEE}
}

@article{pai2023experimentally,
  title={Experimentally realized in situ backpropagation for deep learning in photonic neural networks},
  author={Pai, Sunil and Sun, Zhanghao and Hughes, Tyler W and Park, Taewon and Bartlett, Ben and Williamson, Ian AD and Minkov, Momchil and Milanizadeh, Maziyar and Abebe, Nathnael and Morichetti, Francesco and others},
  journal={Science},
  volume={380},
  number={6643},
  pages={398--404},
  year={2023},
  publisher={American Association for the Advancement of Science}
}

@article{conte2025,
  title={Multimodal Physical Learning in Brain-Inspired Iontronic Networks},
  author={Conte, Monica and van Roij, Ren{\'e} and Dijkstra, Marjolein},
  journal={Preprint at:  	
  https://doi.org/10.48550/arXiv.2511.04209},
  year={2025}
}

@article{Ashtiani2022,
  title = {An on-chip photonic deep neural network for image classification},
  volume = {606},
  ISSN = {1476-4687},
  url = {http://dx.doi.org/10.1038/s41586-022-04714-0},
  DOI = {10.1038/s41586-022-04714-0},
  number = {7914},
  journal = {Nature},
  publisher = {Springer Science and Business Media LLC},
  author = {Ashtiani,  Farshid and Geers,  Alexander J. and Aflatouni,  Firooz},
  year = {2022},
  month = jun,
  pages = {501–506}
}

@article{pai2019matrix,
  title={Matrix optimization on universal unitary photonic devices},
  author={Pai, Sunil and Bartlett, Ben and Solgaard, Olav and Miller, David AB},
  journal={Physical review applied},
  volume={11},
  number={6},
  pages={064044},
  year={2019},
  publisher={APS}
}

@article{williamson2019reprogrammable,
  title={Reprogrammable electro-optic nonlinear activation functions for optical neural networks},
  author={Williamson, Ian AD and Hughes, Tyler W and Minkov, Momchil and Bartlett, Ben and Pai, Sunil and Fan, Shanhui},
  journal={IEEE Journal of Selected Topics in Quantum Electronics},
  volume={26},
  number={1},
  pages={1--12},
  year={2019},
  publisher={IEEE}
}

@article{clements2016optimal,
  title={Optimal design for universal multiport interferometers},
  author={Clements, William R and Humphreys, Peter C and Metcalf, Benjamin J and Kolthammer, W Steven and Walmsley, Ian A},
  journal={Optica},
  volume={3},
  number={12},
  pages={1460--1465},
  year={2016},
  publisher={Optical Society of America}
}

@article{pai2019parallelfaulttolerantprogrammingarbitrary,
  title={Parallel fault-tolerant programming of an arbitrary feedforward photonic network},
  author={Pai, Sunil and Williamson, Ian AD and Hughes, Tyler W and Minkov, Momchil and Solgaard, Olav and Fan, Shanhui and Miller, David AB},
  journal={Preprint at:  	
  https://doi.org/10.48550/arXiv.1909.06179},
  year={2019}
}

@article{Chen2025,
  title = {Self-Contrastive Forward-Forward algorithm},
  volume = {16},
  ISSN = {2041-1723},
  url = {http://dx.doi.org/10.1038/s41467-025-61037-0},
  DOI = {10.1038/s41467-025-61037-0},
  number = {1},
  journal = {Nature Communications},
  publisher = {Springer Science and Business Media LLC},
  author = {Chen,  Xing and Liu,  Dongshu and Laydevant,  Jérémie and Grollier,  Julie},
  year = {2025},
  month = jul 
}

@article{Scodellaro2025,
  title = {Training convolutional neural networks with the Forward–Forward Algorithm},
  volume = {15},
  ISSN = {2045-2322},
  url = {http://dx.doi.org/10.1038/s41598-025-26235-2},
  DOI = {10.1038/s41598-025-26235-2},
  number = {1},
  journal = {Scientific Reports},
  publisher = {Springer Science and Business Media LLC},
  author = {Scodellaro,  Riccardo and Kulkarni,  Ajinkya and Alves,  Frauke and Schr\"{o}ter,  Matthias},
  year = {2025},
  month = nov 
}

@article{Padmani2025,
  title={Function regression using the forward forward training and inferring paradigm},
  author={Padmani, Shivam and Joshi, Akshay},
  journal={Preprint at:  	
  https://doi.org/10.48550/arXiv.2510.06762},
  year={2025}
}

@article{Ashtiani2026,
  title = {Integrated photonic neural network with on-chip backpropagation training},
  ISSN = {1476-4687},
  url = {http://dx.doi.org/10.1038/s41586-026-10262-8},
  DOI = {10.1038/s41586-026-10262-8},
  journal = {Nature},
  publisher = {Springer Science and Business Media LLC},
  author = {Ashtiani,  Farshid and Idjadi,  Mohamad Hossein and Kim,  Kwangwoong},
  year = {2026},
  month = mar 
}

\end{document}

\flushbottom
\maketitle

\large

\section{Generation of prototype vectors}

We generate the prototype vectors in two steps: first, we construct maximally separated vectors using a simplex, and then we randomly orient the simplex.

\textbf{Step 1:} \textit{Generate maximally-separated vectors using a simplex}

Let $c$ denote the number of classes and $d$ the dimensionality of the prototype vectors. The central idea is to use the $c$ vertices of a $(c-1)$-simplex in $d$ dimensions (with $d\geq c$) as the prototype vectors. We first consider the case when $d=c$ and then generalize the approach to $d\geq c$. Note that $d\geq c$ is a constraint for the classification setting due to the one-hot encoded class labels.

A standard $(c-1)$-simplex in $\Rset^c$ by definition is given  by
\be
\Delta^{(c-1)} =\{\bfu\in\Rset^c : \sum_{i=1}^c u_i=1,\ u_i\geq 0 \ \text{for}\ i = 1,\dots,c\}.
\ee
The vertices of $\Delta^{(c-1)}$ can be rearranged as columns of the matrix:
\begin{equation}
\bfX = 
\begin{bmatrix}
    1 & 0 & \cdots & 0 \\
    0 & 1 & \cdots & 0 \\
    \vdots & \vdots & \ddots & \vdots \\
    0 & 0 & \cdots & 1
\end{bmatrix}_{c\times c},
\end{equation}
which is the identity matrix. The centroid of the simplex is $(1/c,1/c,\dots,1/c)^T$. Translating the simplex such that the centroid coincides with the origin yields the modified vertices:
\be
\tilde \bfX =\begin{bmatrix}
    (1 -1/c) & -1/c & \cdots & -1/c \\
    -1/c & (1 -1/c) & \cdots & -1/c \\
    \vdots & \vdots & \ddots & \vdots \\
    -1/c & -1/c & \cdots & (1 -1/c)
\end{bmatrix}_{c\times c}.
\ee
We require the prototype vectors to lie on the sphere $S^{(c-1)}$; therefore, the vertex locations should be normalized so that their distance from the origin equals one. The  distance $z$ of any vertex from the origin is given by
\be
z = \Bigg[\left(1-\frac{1}{c}\right)^2 + \underbrace{\left(-\frac{1}{c}\right)^2 + \left(-\frac{1}{c}\right)^2 + \cdots + \left(-\frac{1}{c}\right)^2}_{(c-1)\ \text{times}}\Bigg]^{1/2} = \left(1 - \frac{1}{c}\right)^{1/2}.
\ee
Therefore, the prototype vectors are given by the columns of the matrix 
\be\label{eq:final_simplex}
\hat\bfX = \frac{1}{z}\begin{bmatrix}
    (1 -1/c) & -1/c & \cdots & -1/c \\
    -1/c & (1 -1/c) & \cdots & -1/c \\
    \vdots & \vdots & \ddots & \vdots \\
    -1/c & -1/c & \cdots & (1 -1/c)
\end{bmatrix}_{c\times c}.
\ee

We will now prove that the prototype vectors constructed from the simplex in equation \eqref{eq:final_simplex} are maximally separated.

For $c \geq 2$, let $\hat{\bfX} = [\hat{\bfx}_1, \hat{\bfx}_2, \dots, \hat{\bfx}_c]$ denote a matrix whose $c$ columns satisfy $\hat{\bfx}_i \in S^{(c-1)}$ for $i=1,\dots,c$. The column vectors are maximally separated if 
\begin{itemize}[itemsep=4pt,topsep=0pt]
    \item $\sum_{i=1}^c \hat\bfx_i= \bm{0}$\quad and
    \item $\hat\bfx_i\cdot\hat\bfx_j = \hat\bfx_i\cdot\hat\bfx_k \quad \forall \ i,j,k=1,\dots,c \quad\text{and}\quad i\neq j\neq k$.
\end{itemize} 
For a detailed proof, see ref.\cite{Kasarla2022} (Theorem 1).

The first condition is trivially satisfied by $\hat\bfX$ in equation \eqref{eq:final_simplex}. For the second condition, the dot product of two distinct  column vectors yields
\be
\begin{aligned}
\hat\bfx_i\cdot\hat\bfx_j &= \frac{1}{z^2}\Bigg[
\left(1-\frac{1}{c}\right)\left(-\frac{1}{c}\right) + \left(1-\frac{1}{c}\right)\left(-\frac{1}{c}\right) + \underbrace{\frac{1}{c^2} \dots +  \frac{1}{c^2}}_{(c-2)\ \text{times}}
\Bigg] \qquad \text{for}\quad i\neq j\\
&= \frac{c}{(c-1)} \Bigg[
\frac{(1-c)}{c^2} + \frac{(1-c)}{c^2} + \frac{(c-2)}{c^2}
\Bigg]\\
% &=- \frac{c}{(c-1)} \frac{c}{c^2}\\
&=-\frac{1}{(c-1)}.
\end{aligned}
\ee
Since the dot product is independent of the indices $i$ and $j$, the second condition is also satisfied. Therefore, the column vectors of $\hat{\bfX}$ are maximally separated prototype vectors.

In the more general case when $d\geq c$, i.e., there are more dimensions than classes, equation \eqref{eq:final_simplex} can simply be extended by appending zeros for the additional $(d-c)$ dimensions. Therefore, the prototype vectors are given by the column vectors of 
\be
\hat \bfX =\frac{1}{z}\begin{bmatrix}
    (1 -1/c) & -1/c & \cdots & -1/c \\
    -1/c & (1 -1/c) & \cdots & -1/c \\
    \vdots & \vdots & \ddots & \vdots \\
    -1/c & -1/c & \cdots & (1 -1/c)\\
    0 & 0 & \cdots & 0\\
    \vdots & \vdots & \ddots & \vdots \\
    0 & 0 & \cdots & 0
\end{bmatrix}_{d\times c} \qquad \text{with}
\quad d\geq c.
\ee

\textbf{Step 2:} \textit{Randomly orient the simplex}

The prototypes generated so far all share the same orientation. However, to avoid any unintentional bias in the learning process, it is beneficial to randomize the orientation of the prototypes, or equivalently, of the simplex. We achieve this by generating a random rotation matrix $\bfT \in \mathrm{SO}(d)$ and rotating each prototype as
\be
\hat\bfx_i \leftarrow \bfT \hat\bfx_i \qquad \text{for} 
\quad i=1,\dots,c.
\ee

To generate rotation matrices from a uniform distribution on $\text{SO}(d)$, we adapt the algorithm of ref.\cite{Mezzadri2006} (Section 5) for generating unitary matrices. Specifically, we implement a real-valued analog of the algorithm from ref.~\cite{Mezzadri2006} to sample orthogonal matrices, followed by a rejection step for matrices with negative determinant, thereby yielding
random special orthogonal matrices, i.e., rotation matrices. The proposed algorithm for generating random rotation matrices is as follows:
\begin{itemize}[itemsep=4pt,topsep=0pt]
\item Sample $\bfM \in \Rset^{d\times d}$ via $\bfA\sim \calN(\bm{0},\bfI)$.
\item Perform QR decomposition of $\bfM = \bfQ\bfR$, where $\bfQ$ is an orthogonal matrix and $\bfR$ is an upper-triangular matrix.
\item Construct the diagonal matrix $\bfA = \begin{bmatrix}
    \frac{R_{11}}{|R_{11}|} & 0 & \cdots & 0 \\
    0 & \frac{R_{22}}{|R_{22}|} & \cdots & 0 \\
    \vdots & \vdots & \ddots & \vdots \\
    0 & 0 & \cdots & \frac{R_{dd}}{|R_{dd}|}
\end{bmatrix}_{d\times d}$.
\item If $\det(\bfQ\bfA)<0$, restart from sampling $\bfM$.
\item Return $\bfT = \bfQ\bfA$.
\end{itemize}

\section{Expectation of directional-derivative-based weights update}

Given the random directional perturbations applied to the weights, we now compute the expected value of the resulting weight update. 
Recall that the directional-derivative-based update is given by
\be
\bfomega  \leftarrow \bfomega + \lambda \underbrace{  n 
\nabla_{\bfv}\calG(\bfomega)
\frac{\bfv}{\|\bfv\|}}_{\text{weights update}}
\qquad \text{with} \quad 
\bfv\sim\calN(\boldsymbol{0},\bfI) 
\quad \text{and} \quad
n=\dim(\bfomega) =\dim(\bfv).
\ee
The directional derivative is defined as
\be\label{eq:dd_def}
\nabla_{\bfv}\calG(\bfomega) = \partderiv{\calG(\bfomega)}{\bfomega}\cdot\frac{\bfv}{\|\bfv\|} = \frac{\bfv^T}{\|\bfv\|}  \partderiv{\calG(\bfomega)}{\bfomega}
\ee

Consider the expectation of the weights update over $\bfv$ and substitute equation \eqref{eq:dd_def}:
\be\label{eq:dd_update_expectation_intermediate}
\mathbb{E}_{\bfv}\left[n \nabla_{\bfv}\calG(\bfomega) \frac{\bfv}{\|\bfv\|} \right] 
=n\ \mathbb{E}_{\bfv}\left[
\frac{\bfv}{\|\bfv\|}\left(\frac{\bfv^T}{\|\bfv\|}  \partderiv{\calG(\bfomega)}{\bfomega}\right)
 \right]
=n\ \mathbb{E}_{\bfv}\left[
\frac{\bfv\bfv^T}{\|\bfv\|^2} 
 \right]
\partderiv{\calG(\bfomega)}{\bfomega}.
\ee
The trace of the remaining expectation term $\mathbb{E}_{\bfv}[\bfv\bfv^T/\|\bfv\|^2]$ simplifies as:
\be\label{eq:tr1}
\tr \left(\mathbb{E}_{\bfv}\left[
\frac{\bfv\bfv^T}{\|\bfv\|^2} 
\right]\right) 
= \mathbb{E}_{\bfv}\left[\tr\left(
\frac{\bfv\bfv^T}{\|\bfv\|^2} \right)
 \right]
= \mathbb{E}_{\bfv}\left[
\sum_{j=1}^n\frac{v_j^2}{\|\bfv\|^2} 
\right]
= \mathbb{E}_{\bfv}\left[
\frac{\|\bfv\|^2}{\|\bfv\|^2} 
\right]
= 1.
\ee
Additionally, since the probability distribution of $\bfv$ is isotropic,
\be\label{eq:tr2}
\mathbb{E}_{\bfv}\left[
\sum_{j=1}^n\frac{v_j^2}{\|\bfv\|^2} 
\right] 
= n\ \mathbb{E}_{\bfv}\left[
 \frac{v_i^2}{\|\bfv\|^2} 
\right]
\quad \forall\ i=1,\dots,n.
\ee
Comparing equation \eqref{eq:tr1} and equation  \eqref{eq:tr2} yields the diagonal terms of the expectation:
\be\label{eq:diagonal}
\mathbb{E}_{\bfv}\left[
 \frac{v_i^2}{\|\bfv\|^2} \right] = \frac{1}{n}\quad  \forall\ i=1,\dots,n.
\ee
Now, consider the off-diagonal terms of the expectation:
\be
\mathbb{E}_{\bfv}\left[
 \frac{v_i v_j}{\|\bfv\|^2} \right] 
 = \mathbb{E}_{\bfv}\left[
 \frac{v_i (-v_j)}{\|\bfv\|^2} \right]
 = - \mathbb{E}_{\bfv}\left[
 \frac{v_i v_j}{\|\bfv\|^2} \right]
 \quad\forall \ i,j=1,\dots,n \ \ \text{and}\ \ i\neq j,
\ee
where the symmetry of the distribution of $\bfv$ is used to obtain the middle term. Simplifying further yields: 
\be\label{eq:off-diagonal}
\mathbb{E}_{\bfv}\left[
 \frac{v_i v_j}{\|\bfv\|^2} \right] 
 = 0
 \quad\forall \ i,j=1,\dots,n \ \ \text{and}\ \ i\neq j,
\ee
Combining equation \eqref{eq:diagonal} and equation \eqref{eq:off-diagonal} yields
\be\label{eq:vvt}
\mathbb{E}_{\bfv}\left[
 \frac{\bfv \bfv^T}{\|\bfv\|^2} \right] = \frac{1}{n}\bfI.
\ee
Substituting equation \eqref{eq:vvt} into equation \eqref{eq:dd_update_expectation_intermediate} gives the expectation of the directional-derivative-based update as
\be
\mathbb{E}_{\bfv}\left[n \nabla_{\bfv}\calG(\bfomega) \frac{\bfv}{\|\bfv\|} \right] 
=n\ \frac{1}{n}\bfI
\partderiv{\calG(\bfomega)}{\bfomega},
\ee
which simplifies to
\be
\mathbb{E}_{\bfv}\left[n \nabla_{\bfv}\calG(\bfomega) \frac{\bfv}{\|\bfv\|} \right] 
=\partderiv{\calG(\bfomega)}{\bfomega}.
\ee
Therefore, the expectation of the directional-derivative-based update equals the partial-derivative-based update.

\section{Influence of the number of directions used in estimation of directional derivatives}

\begin{figure}
    \centering
\includegraphics[width=\linewidth]{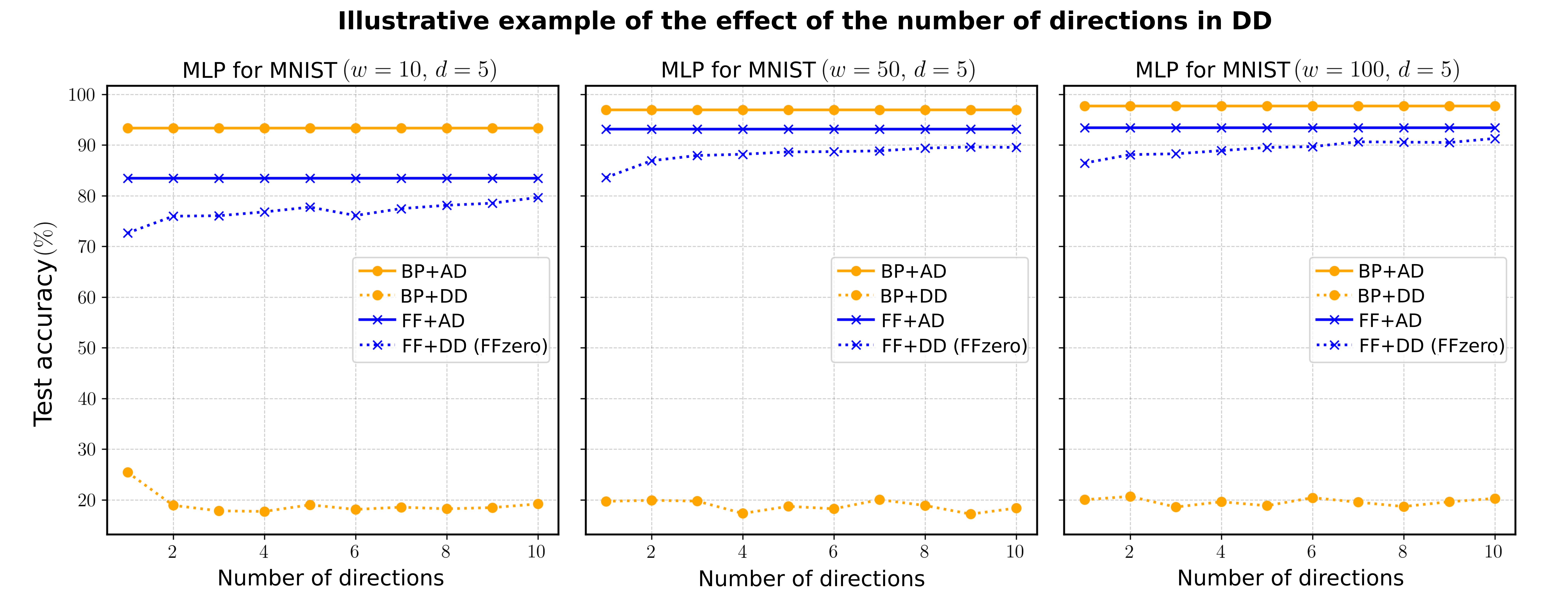}
\caption{Classification accuracy of MLP models across varying number of sampled directions used in directional-derivative (DD) optimization.  Results are shown for MNIST dataset with MLPs of fixed depth $d = 5$ and different hidden-layer widths: $w \in \{10, 50, 100\}$, with all hidden layers constrained to equal size.  The results are compared between backpropagation (BP) and forward–forward (FF) training paradigms, each combined with either automatic differentiation (AD) or directional-derivative (DD) optimization. }
\label{fig:direction}
\end{figure}

In \figurename~\ref{fig:direction}, we study the impact of the number of sampled directions ($P$) used in directional-derivative (DD) estimation on optimization performance and generalization. Multilayer perceptrons (MLPs) are trained on the MNIST dataset with a fixed depth of $d = 5$ and varying hidden-layer widths $w \in \{10, 50, 100\}$. Test accuracy is evaluated as a function of the number of directions used to approximate directional information during optimization.

Across all model sizes, methods based on FF+DD show a clear improvement in test accuracy as the number of sampled directions increases. The gains are most significant for small numbers of directions  and gradually saturate as more directions are added, indicating diminishing returns beyond a moderate 3-4 number of sampled directions. In contrast, BP+DD exhibits substantially weaker sensitivity to the number of directions and consistently lower accuracy. These results indicate that DD-based FF estimation exploits directional information more effectively than DD-based BP. The observed trends are consistent across network widths, suggesting that the advantage of FFzero is robust with respect to model complexity.

\section{Layer embedding of photonic neural network trained on MNIST using FFzero}

\begin{figure}[t]
\centering
\includegraphics[width=\linewidth]{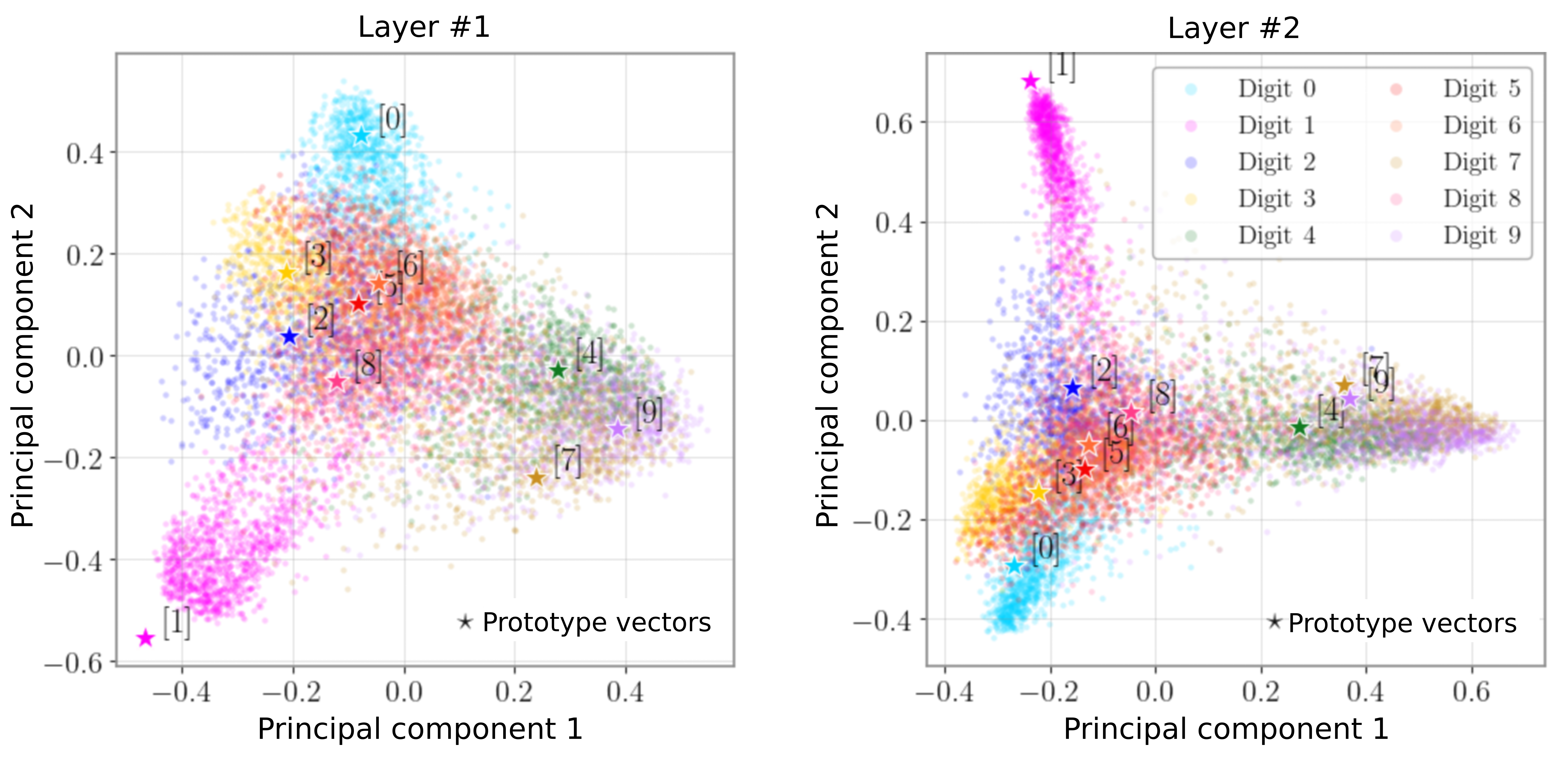}
\caption{
PCA projection of the layer-wise
embeddings of a two-layer photonic neural network trained with FFzero on MNIST. Each point represents one test sample, colored by the digit class. Stars denote the pre-computed prototype vectors projected into the same PCA space. After training with FFzero, data points cluster towards their corresponding prototype vectors in all layers.
}
\label{fig:pnn_pca}
\end{figure}

We perform Principal Component Analysis (PCA)  on the layer-wise intensity
outputs of the photonic neural network after training on MNIST with FFzero. The PCA of the resulting outputs, together with the pre-computed prototype vectors, are shown in Fig.~\ref{fig:pnn_pca} for each layer.
In both layers, the embeddings of the same class cluster around their corresponding prototype vector, confirming that the cosine-similarity objective successfully aligns the intensity outputs with the prototype vectors.

\section{Implementation details}

\subsection{Architecture of MLPs in FFzero benchmarks}

\begin{figure}
\includegraphics[width=\linewidth]{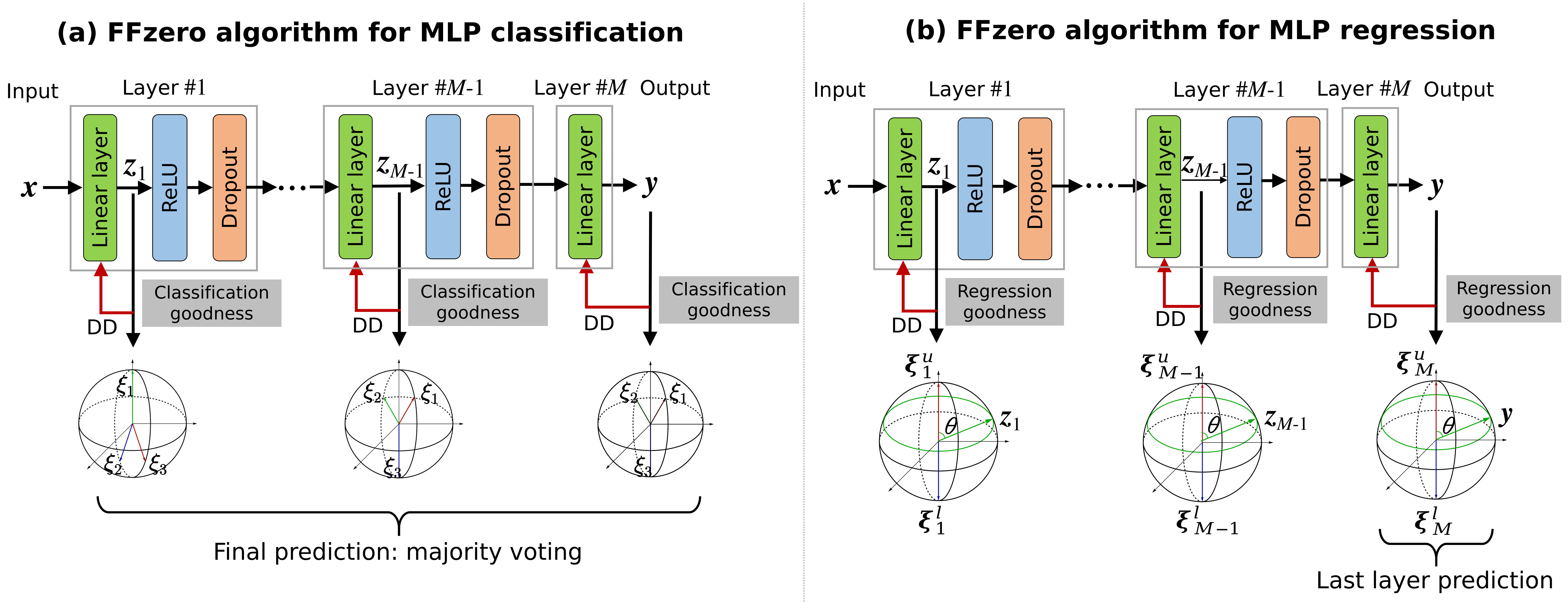}
\caption{Schematic illustration of the FFzero algorithm and architecture used  for MLPs in (\textbf{a}) classification and (\textbf{b}) regression tasks.}
\label{fig:mlp-implementation}
\end{figure}

\figurename~\ref{fig:mlp-implementation} schematically illustrates the architecture of the MLPs used in the FFzero benchmarks.

\subsection{Architecture of CNNs in FFzero benchmarks}

\begin{figure}
\includegraphics[width=\linewidth]{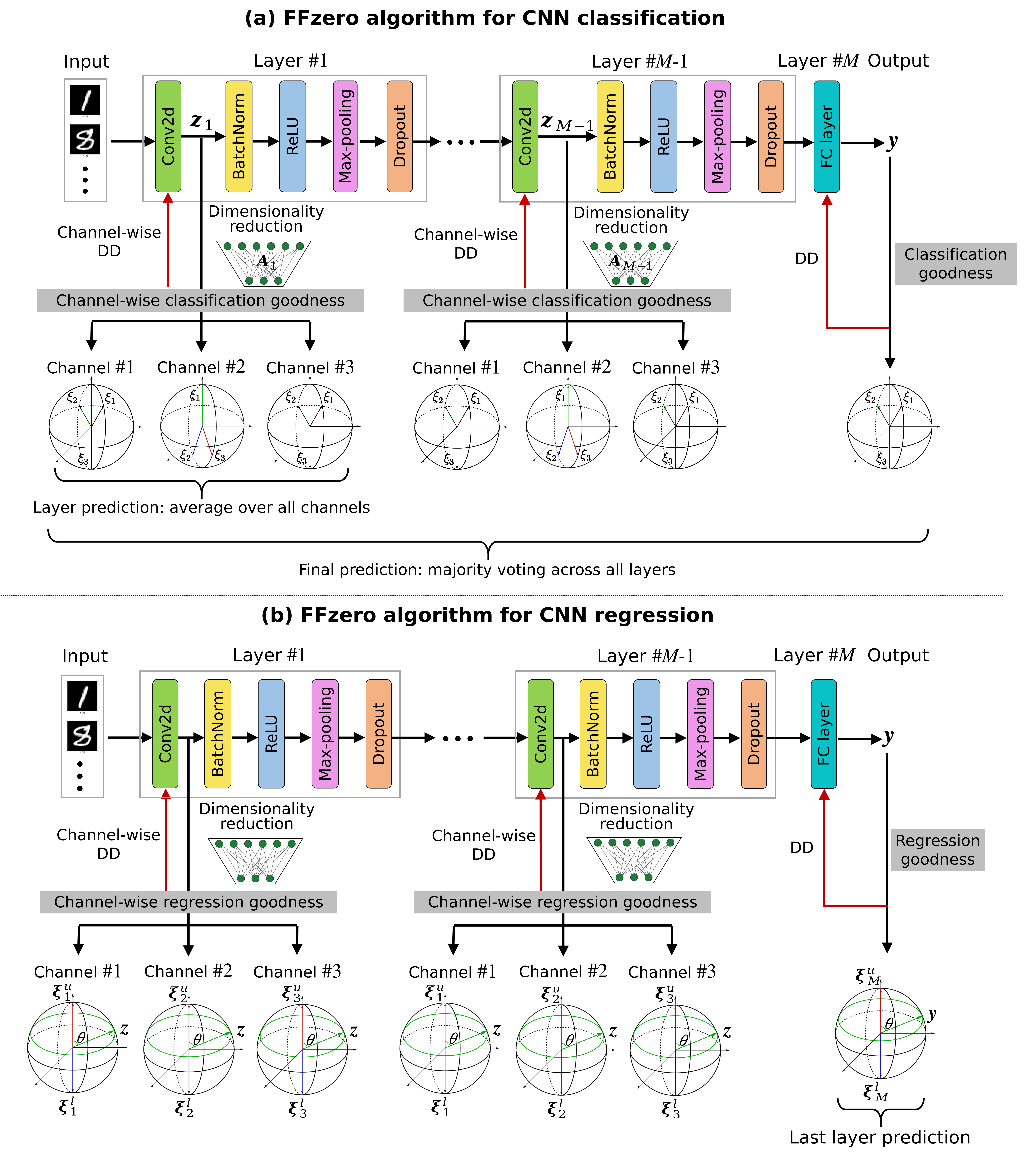}
\caption{Schematic illustration of the FFzero algorithm and architecture used  for CNNs in (\textbf{a}) classification and (\textbf{b}) regression tasks. Within each convolution layer, feature maps from individual channels are reshaped and projected to a lower-dimensional space via fixed random linear mappings. The reduced representations are compared with fixed prototype vectors using cosine similarity, and a channel-wise goodness is computed to update the corresponding channel parameters through DD.}
\label{fig:cnn-implementation}
\end{figure}

\figurename~\ref{fig:cnn-implementation} schematically illustrates the architecture of the CNNs used in the FFzero benchmarks. In the following, we elaborate upon the dimensionality reduction and training strategy for CNNs in FFzero.

Let the input to a convolutional layer be $\bfx \in \mathbb{R}^{C_{\mathrm{in}} \times H_{\mathrm{in}} \times W_{\mathrm{in}}}$,
where $C_{\mathrm{in}}$ denotes the number of input channels.
Consider a convolutional layer with $C_{\mathrm{out}}$ filters, kernel size
$K_h \times K_w$, stride $S_h \times S_w$, padding $P_h \times P_w$,
and dilation $D_h \times D_w$.
Then, the output of the layer is 
$
\bfz \in \mathbb{R}^{C_{\mathrm{out}} \times H_{\mathrm{out}} \times W_{\mathrm{out}}}
$, 
where the spatial dimensions are given by 
\be
H_{\mathrm{out}} =
\left\lfloor
\frac{H_{\mathrm{in}} + 2P_h - D_h (K_h - 1) - 1}{S_h} + 1
\right\rfloor,
\qquad
W_{\mathrm{out}} =
\left\lfloor
\frac{W_{\mathrm{in}} + 2P_w - D_w (K_w - 1) - 1}{S_w} + 1
\right\rfloor.
\ee
The output of the $i^\text{th}$ channel, with spatial size
\( H_{\mathrm{out}} \times W_{\mathrm{out}} \), is then reshaped into a vector
$\bfu^{(i)} \in \mathbb{R}^{H_{\mathrm{out}} W_{\mathrm{out}}}$
and passed through a dimensionality reduction layer with fixed, randomly initialized, and non-trainable parameters,
\be
\tilde{\bfz}^{(i)} = \bfA^{(i)}\bfu^{(i)},
\ee
where $\bfA^{(i)} \in \mathbb{R}^{d \times H_{\mathrm{out}} W_{\mathrm{out}}} \) and \( d \ll H_{\mathrm{out}} W_{\mathrm{out}}$. 
A channel-wise goodness metric is then evaluated based on the low-dimensional representation $\tilde{\bfz}$,
and channel-wise directional derivatives are used to update the parameters of each channel locally and sequentially.

In each convolution layer, predictions are obtained from channel-wise low-dimensional representations. For classification tasks, the prediction is obtained by computing a channel-wise goodness score based on the similarity between the reduced feature representations and the corresponding prototype vectors. These goodness scores are accumulated across all output channels, and the class associated with the maximum aggregated goodness is selected as the layer-wide prediction.
For regression tasks, the prediction is computed by averaging the similarity scores over all output channels. In both cases, feature representations are normalized prior to similarity evaluation, ensuring that the goodness measure corresponds to cosine similarity.

\subsection{Data generation for synthetic function regression tasks}

Synthetic regression datasets are generated from prescribed analytical functions (see (15) in main article) for benchmarking. 
For each case, a total of 12{,}000 input samples are drawn independently from a uniform distribution on $[-1,1]^d$, where $d$ denotes the input dimension. The dataset is randomly permuted and split into 10{,}000 training samples and 2{,}000 test samples. To emulate measurement noise, zero-mean Gaussian noise with a standard deviation equal to $5\%$ of the standard deviation of the clean training outputs is added to the training targets only. The outputs are subsequently normalized to the interval $[-1,1]$ according to
\[
\tilde{y} = 2\,\frac{y - y_{\min}}{y_{\max} - y_{\min}} - 1,
\]
where $y_{\min}$ and $y_{\max}$ denote the minimum and maximum values of the noisy training targets, respectively. The same normalization is applied to the test targets. See Supplementary Table~\ref{tab:parameters} for the parameters used in dataset generation and the model architecture.

\subsection{Parameters and hyperparameters used in experiments.}

The parameters and hyperparameters used in the classification and regression tasks are listed in Supplementary Table~\ref{tab:parameters}.

\begin{table}[t]
\setcounter{table}{0}
\centering
\caption{List of parameters and hyperparameters used in experiments.} 
\label{tab:parameters}
\begin{tabular}{lcc}
\hline
\multicolumn{1}{l}{\textbf{Parameter}} & \textbf{Notation} & \textbf{Value} \\ \hline
\textit{MLP for MNIST classification, MNIST regression, $\qquad\qquad\qquad\quad$} \\
\textit{and FashionMNIST classification:}\\
$\quad$ Size of training dataset  & - & $50000$ \\
$\quad$ Size of test dataset   & -  & $10000$ \\ 
$\quad$ Batch size & - & $256$  \\
$\quad$ Number of training epochs  & - & $100$  \\
$\quad$ Learning rate  & $\lambda$ & $0.001$  \\
$\quad$ Step size in DD  & $\epsilon$ & $0.001$  \\
$\quad$ Number of directions in DD  & $P$ & $1$  \\
$\quad$ Input dimension  & - & $784$  \\
$\quad$ Output dimension  & - & $10^{*}\text{ or }1^{**}$  \\
$\quad$ Hidden layer dimension  & - & $\{10, 50, 100\}$  \\
$\quad$ Number of hidden layers  & - & $\{1, 2, 3, 4, 5, 6, 7, 8, 9, 10\}$  \\
$\quad$ Dropout & - & $0$  \\
$\quad$ Activation function  & - & ReLU  \\
$\quad$ Margin loss in classification tasks & $q$ & 0.3 \\
$\quad$ Loss function in BP & - & Cross-entropy loss$^{*}$ or MSE loss$^{**}$ \\
\hline
\textit{CNN for MNIST classification and regression:}\\
$\quad$ Size of training dataset  & - & $50000$ \\
$\quad$ Size of test dataset   & -  & $10000$ \\ 
$\quad$ Batch size & - & $256$  \\
$\quad$ Number of training epochs  & - & $100$  \\
$\quad$ Learning rate  & $\lambda$ & $0.001$  \\
$\quad$ Step size in DD  & $\epsilon$ & $0.001$  \\
$\quad$ Number of directions in DD  & $P$ & $1$  \\
$\quad$ Input dimension  & - & $28\times28$  \\
$\quad$ Output dimension  & - & $10^{*}\text{ or }1^{**}$  \\
$\quad$ Number of hidden convolution layers  & - & $2$  \\
$\quad$ Number of channels in each hidden layer & - & $\{4, 8, 12, 16, 20, 24, 28, 32\}$  \\
$\quad$ Padding size & - & $2$  \\
$\quad$ Kernel size of each channel & - & $6\times6$  \\
$\quad$ Kernel stride & - & $1$  \\
$\quad$ Pooling size & - & $2\times2$  \\
$\quad$ Pooling stride & - & $2$  \\
$\quad$ Dropout & - & $0$  \\
$\quad$ Dimension of prototype vectors of each channel & - & $10$  \\
$\quad$ Output dimension of fully connected layer & - & $10^{*}\text{ or }1^{**}$  \\
$\quad$ Dropout of fully connected layer & - & $0$  \\
$\quad$ Activation function  & - & ReLU  \\
$\quad$ Margin loss in classification tasks & $q$ & 0.3 \\
$\quad$ Loss function in BP & - & Cross-entropy loss$^{*}$ or MSE loss$^{**}$ \\
\hline
\end{tabular}
\begin{tablenotes}
  \item $^*$ Classification tasks.
  \item $^{**}$ Regression tasks.
\end{tablenotes}
\end{table}

\begin{table}[t]
\begin{center}
\setcounter{table}{0}
\caption{(continued) List of parameters and hyperparameters used in experiments.} 
\label{tab:parameters}
\begin{tabular}{lcc}
\hline
\multicolumn{1}{l}{\textbf{Parameter}} & \textbf{Notation} & \textbf{Value} \\ 
\hline
\textit{CNN for FashionMNIST classification:}\\
$\quad$ Size of training dataset  & - & $50000$ \\
$\quad$ Size of test dataset   & -  & $10000$ \\ 
$\quad$ Number of training epochs  & - & $100$  \\
$\quad$ Learning rate  & $\lambda$ & $0.001$  \\
$\quad$ Step size in DD  & $\epsilon$ & $0.001$  \\
$\quad$ Number of directions in DD  & $P$ & $1$  \\
$\quad$ Input dimension  & - & $28\times28$  \\
$\quad$ Output dimension  & - & $10$  \\
$\quad$ Number of hidden convolution layers  & - & $2$  \\
$\quad$ Number of channels in each hidden layer & - & $\{8, 16, 24, 32, 40, 48, 56, 64\}$  \\
$\quad$ Padding size & - & $2$  \\
$\quad$ Kernel size of each channel & - & $6\times6$  \\
$\quad$ Kernel stride & - & $1$  \\
$\quad$ Pooling size & - & $2\times2$  \\
$\quad$ Pooling stride & - & $2$  \\
$\quad$ Dropout of each convolution layer & - & $0.1$  \\
$\quad$ Dimension of prototype vectors of each channel & - & $10$  \\
$\quad$ Output dimension of fully connected layers & - & $\{100, 10\}$  \\
$\quad$ Dropout of fully connected layers & - & $\{0.1, 0.\}$  \\
$\quad$ Activation function  & - & ReLU  \\
$\quad$ Margin loss & $q$ & 0.3 \\
$\quad$ Loss function in BP & - & Cross-entropy loss \\
\hline
\textit{MLP for synthetic function regression tasks:}\\
$\quad$ Size of training dataset  & - & $10000$ \\
$\quad$ Size of test dataset   & -  & $2000$ \\ 
$\quad$ Batch size & - & $256$  \\
$\quad$ Number of training epochs  & - & $100$  \\
$\quad$ Learning rate  & $\lambda$ & $0.001$  \\
$\quad$ Step size in DD  & $\epsilon$ & $0.001$  \\
$\quad$ Number of directions in DD  & $P$ & $1$  \\
$\quad$ Input range & - & $[-1, 1]$  \\
$\quad$ Input dimension  & - & $2^{a}\text{ or }5^{b}$  \\
$\quad$ Output dimension  & - & $1$  \\
$\quad$ Number of hidden layers  & - & $2$  \\
\multirow{3}{15em}{$\quad$ Hidden layer dimension}  &  & $\{10, 20, 30, 40, 50, 60, 70, 80, 90, 100\}^{a}$  \\
& - & \text{ or }  \\
&  & $\{20, 40, 60, 80, 100, 120, 140, 160, 180, 200\}^{b}$  \\
$\quad$ Loss function in BP & - & MSE loss \\
\hline
\textit{Photonic emulation:}\\
$\quad$ Size of training dataset  & - & $60000$ \\
$\quad$ Size of test dataset   & -  & $10000$ \\ 
$\quad$ Batch size & - & $128$  \\
$\quad$ Number of training epochs  & - & $100$  \\
$\quad$ Learning rate  & $\lambda$ & $0.001$  \\
$\quad$ Step size in DD  & $\epsilon$ & $0.001$  \\
$\quad$ Number of directions in DD  & $P$ & $1$  \\
$\quad$ MZI dimension  & - & $784\times784$  \\
$\quad$ Number of MZI layers  & - & $2$  \\
$\quad$ Loss function in FF & - & Equation~(5) in main article \\
$\quad$ Loss function in BP & - & Cross-entropy loss \\
\hline
\end{tabular}
\end{center}
\begin{tablenotes}
  \item $^a$ Function 1.
  \item $^b$ Function 2.
\end{tablenotes}
\end{table}

\bibliography{Bib}